\documentclass{article}

\usepackage{microtype}
\usepackage{graphicx}
\usepackage{subcaption}
\usepackage{tabularx}  
\usepackage{booktabs} 

\usepackage{hyperref}

 \usepackage[preprint]{icml2026}

\usepackage{amsmath}
\usepackage{amssymb}
\usepackage{mathtools}
\usepackage{amsthm}

\usepackage[capitalize,noabbrev]{cleveref}

\theoremstyle{plain}

\theoremstyle{definition}

\theoremstyle{remark}

\usepackage[textsize=tiny]{todonotes}

\icmltitlerunning{K-U-KAN: Koopman-Enhanced U-KAN  for 3D Dental Reconstruction from a Single Panoramic X-ray Radiograph}

\begin{document}

\twocolumn[
  \icmltitle{K-U-KAN: Koopman-Enhanced U-KAN  for 3D Dental Reconstruction from a Single Panoramic X-ray Radiograph}



  \icmlsetsymbol{equal}{*}

  \begin{icmlauthorlist}
    \icmlauthor{Bikram Keshari Parida}{yyy,yy}
    \icmlauthor{Abhijit Sen}{yy}
    \icmlauthor{Wonsang You}{yyy}
  \end{icmlauthorlist}

  \icmlaffiliation{yyy}{Artificial Intelligence \& Image Processing Lab., Department of Information \& Communication Engineering, Sun Moon University, Asan-Si, South Korea}
  \icmlaffiliation{yy}{Department of Physics and Engineering Physics, Tulane University, New Orleans, LA, USA}

  \icmlcorrespondingauthor{Bikram Keshari Parida}{parida.bikram90.bkp@gmail.com}
  \icmlcorrespondingauthor{Abhijit Sen}{asen1@tulane.edu}
  \icmlcorrespondingauthor{Wonsang You}{wyou@sunmoon.ac.kr}

  \icmlkeywords{Kolmogorov-Arnold Networks (KAN), Koopman Operator, Panoramic X-Ray Radiography, CBCT, 3D reconstruction}

  \vskip 0.3 in
]



\printAffiliationsAndNotice{}  

\begin{abstract}
A panoramic X-ray compresses a 3D jaw into a 2D strip; we aim to recover the missing depth cleanly and fast. Existing implicit neural representations render realistic volumes but are slow to train, sensitive to sampling and positional encodings, and costly in practice. Pure CNN baselines are efficient yet struggle with the dental arch’s long-range geometry, blur fine enamel–dentin boundaries, and offer little interpretability. We present K-U-KAN, a three-stage pipeline that (i) lifts 2D features into depth-aware observables with Kolmogorov–Arnold Networks, (ii) advances these observables by a stable, phase-aware linear evolution via a Koopman token block, and (iii) places the predicted depth bins onto focal-trough rays before a lightweight 3D attention U-KAN refines the volume. This marriage of physics (Beer–Lambert image formation), geometry (horseshoe focal trough), and learned linear dynamics yields sharp anatomy, fewer artifacts, and robust behavior on native radiographic intensities with batch size one. On held-out data, K-U-KAN matches transformer/implicit baselines on signal and structure metrics, clearly improves perceptual quality, and trains in roughly half the time—making single-view PX $\to$ CBCT reconstruction more practical for clinical pipelines.
\end{abstract}

\section{Introduction}

\begin{figure*}[th]
	\centering
	\includegraphics[width=\linewidth]{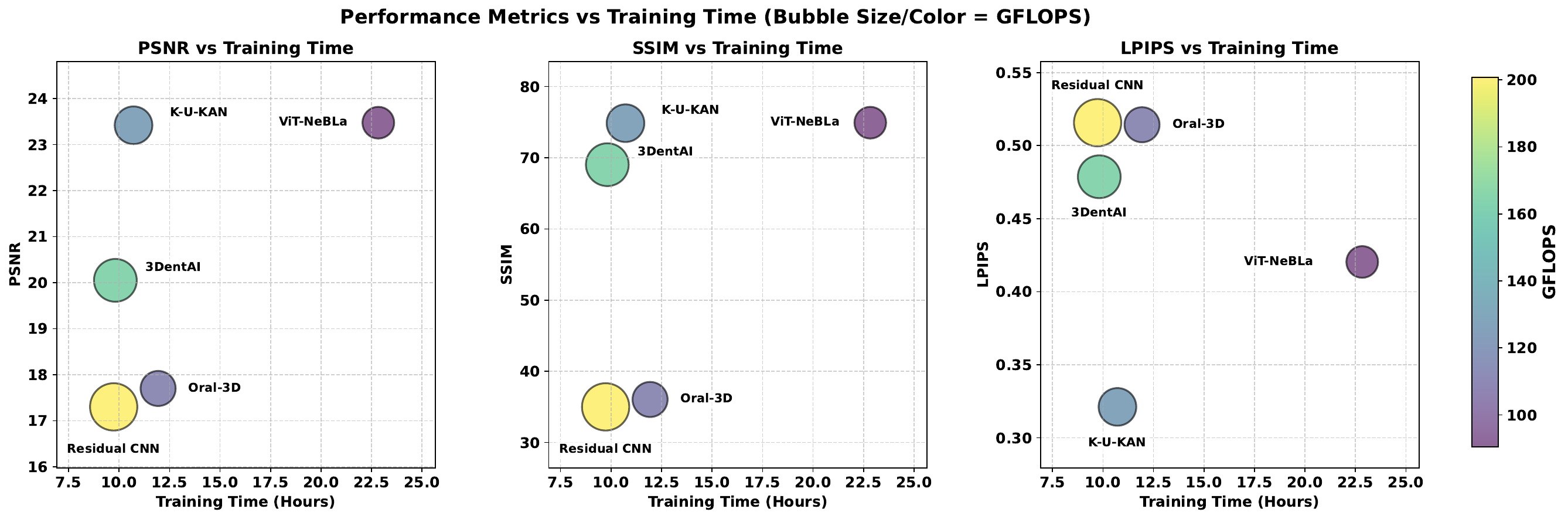}
	\caption{\textbf{Compute–performance trade-off of 3D oral reconstruction models.}
		Bubble plots of PSNR, SSIM, and LPIPS versus wall-clock training time for all compared methods.
		Each bubble corresponds to one model, with its area and color proportional to the per-forward GFLOPS reported in Table~\ref{tab:complexity} (larger and darker bubbles indicate higher computational cost), while metric values and training times are taken from Table~\ref{tab1} and averaged over 10 runs.
		The plots show that \textbf{K-U-KAN} attains ViT-NeBLa–level PSNR/SSIM with a lower LPIPS and substantially shorter training time, whereas CNN-based baselines incur comparable or higher computational cost without reaching the same reconstruction quality, indicating that \textbf{K-U-KAN} offers the most favorable balance between accuracy and efficiency.}
	
	\label{fig:bubble_plots}
\end{figure*}

Reconstructing three--dimensional (3D) maxillofacial anatomy from radiographic data is, at its core, an ill--posed inverse problem. A single projection collapses depth, so that many distinct 3D configurations of teeth and bone can produce similar two--dimensional (2D) images once integrated along the X--ray path \cite{x2teeth,s_deep_2023}. In panoramic imaging, the situation is further complicated by the acquisition geometry: the X--ray source and detector follow an approximately elliptical trajectory designed to keep the dental arch within a focal trough, which induces spatially varying magnification, blurring, and geometric distortion that depend on patient anatomy and positioning \cite{stramotas_accuracy_2002,gribel_accuracy_2011}. Because this forward model is non--linear, only partially known, and strongly anatomy--dependent, analytic inversion is not tractable, and meaningful reconstruction requires additional anatomical priors or data--driven regularization to constrain the solution space.

Within this context, panoramic radiographs (PX) occupy a central place in routine dental practice. PX provides a single global view of the maxillofacial region at low financial cost and relatively low radiation dose, which explains its widespread use as a first--line imaging modality \cite{vandenberghe_2010,shah_recent_2014}. However, the compression of 3D anatomy into a 2D projection inevitably obscures clinically important spatial relationships. Root morphology, the course of neurovascular canals, and the thickness and continuity of cortical bone can often only be inferred indirectly, limiting the reliability of PX for applications that demand precise 3D localization and measurement, such as implant planning, surgical extraction of impacted teeth, and orthodontic treatment planning \cite{scarfe2006,angelopoulos_2012}.

Cone--beam computed tomography (CBCT) was introduced precisely to overcome these limitations by explicitly reconstructing the 3D distribution of X--ray attenuation. By acquiring multiple projections around the patient and applying tomographic reconstruction algorithms, CBCT yields volumetric datasets with high spatial resolution and accurate geometric representation of hard tissues \cite{gupta_cone_2013,jacobs_cone_2018}. For many diagnostic and interventional tasks, CBCT is therefore regarded as the reference standard. Nevertheless, the substantially higher radiation burden and increased cost relative to PX impose constraints under the ALARA (As Low As Reasonably Achievable) principle and preclude its use as a universal screening tool in everyday dentistry \cite{ludlow_2015,oenning_2018}. As a result, clinicians often face a trade--off between the accessibility of PX and the superior 3D information provided by CBCT.

Deep learning methods have recently been explored as a way to bridge this gap by learning effective priors directly from data. Early studies trained convolutional neural networks (CNNs) on paired PX--CBCT datasets to regress volumetric occupancy grids or per--voxel depth maps from a single PX, demonstrating that statistical relationships between 2D image patterns and underlying 3D structures can be captured by suitably designed architectures \cite{x2teeth,yaqub_deep_2022}. Other approaches employed generative adversarial networks (GANs) to synthesize flattened 3D volumes from PX and then deform these volumes along a dental arch curve extracted from the image, thereby approximating the natural curvature of the jaws \cite{song_2021,liang_2021a}. While these methods have shown promising results, they typically focus on restricted anatomical regions---most commonly the dentition---and often rely on tightly paired CBCT data or explicit arch templates, which may not generalize across heterogeneous patient populations and acquisition settings.

These considerations collectively motivate the search for PX--to--3D reconstruction strategies that recover a more comprehensive representation of maxillofacial anatomy while relaxing dependence on densely paired volumetric labels and hand--crafted geometric assumptions. Incorporating explicit models of panoramic projection geometry, quantifying predictive uncertainty, and leveraging weakly supervised or unpaired CBCT data are potential avenues to narrow the gap between conventional PX and CBCT, with the longer--term goal of enabling 3D--aware diagnostic and treatment--planning tools within the constraints of routine dental practice.

In parallel, implicit neural representations have demonstrated that continuous volumetric fields can be learned directly from image data. Methods such as \emph{NeRF} \cite{mildenhall_nerf_2022} and neural attenuation fields (NAF) \cite{naf_2022} recover density or radiance functions from sparse views, and have been adapted for medical imaging scenarios \cite{nerf_survey,NeAS,greenspan_learning_2023,molaei_implicit_2023,snaf}. However, these approaches typically assume access to multiple projections of the same subject and tend to generalize poorly across patients, which limits their applicability in settings where only a single PX is available.

Concurrently, Vision Transformers (ViTs) have reshaped 2D image analysis by using self--attention to model long--range relationships and global context that conventional CNNs capture only indirectly \cite{dosovitskiy_image_2021,khan_transformers_2022}. This suggests a natural synergy: ViTs can extract holistic anatomical cues and global configuration information from a single PX, while an implicit volumetric decoder can map these features into a continuous 3D representation of maxillofacial anatomy \cite{vit_nebla_2025}.

Implicit neural representation (INR) models have demonstrated strong performance for high-resolution 3D dental reconstruction, but their practical deployment is limited by the substantial computational cost associated with dense point-wise evaluations and volumetric rendering. To retain INR-like predictive capability while reducing training and inference time, we integrate Koopman-operator–based dynamical modeling and tokenized Kolmogorov–Arnold Network (KAN) blocks into a conventional U-Net backbone. Conventional U-Nets have a limited ability to capture complex cross-channel nonlinear patterns and offer poor interpretability of what each channel represents, whereas tokenized KAN blocks explicitly model these nonlinear interactions in a more interpretable functional basis. Incorporating a Koopman operator model on top of KAN features further imposes a structured, approximately linear evolution in a lifted latent space, improving stability, data efficiency, and extrapolative behavior compared to using KANs alone. The resulting Koopman–KAN–CNN hybrids approximate the expressive power of INR-based approaches while operating on grid-based representations with hardware-optimized convolutions, thereby substantially reducing computational overhead for fast and accurate 3D dental reconstruction. Fig. \ref{fig:bubble_plots} illustrates the compute-performance trade-off of the 3D oral reconstruction models.

The rest of the paper is organized as follows: Section \ref{sec:rel_work} reviews related work on PX-conditioned 3D dental reconstruction, spanning tooth-level, template/arch-based, and physics-driven implicit approaches, and situates their limitations. Section \ref{sec:background} introduces the theoretical foundations of the proposed framework, including Koopman operator learning and Kolmogorov–Arnold Networks within the U-KAN paradigm. Section \ref{sec:method} details the proposed K-U-KAN methodology, describing the lifting network, Koopman-enhanced token dynamics, focal-trough inverse warping, and the 3D attention-based refinement module. Section \ref{sec:expt} presents the experimental setup, datasets, training protocol, and comprehensive quantitative and qualitative evaluations against state-of-the-art baselines. Finally, Section \ref{sec:conclusion} discusses the implications, limitations, and potential clinical impact of the proposed approach, followed by concluding remarks and future directions.

\section{Related Works} \label{sec:rel_work}

\subsection{Tooth-Level PX-Conditioned Approaches}

A large part of the literature on PX-driven 3D modelling has concentrated on the dentition rather than the entire maxillofacial complex.  Early CNN-based systems first demonstrated that panoramic radiographs contain sufficient information for reliable two-dimensional analysis, achieving high performance in tasks such as tooth localization and detection of caries or periodontal lesions \cite{jader_deep_2018,lee_diagnosis_2018}.  These works operated purely in image space, but they established PX as a rich input modality for data-driven models.

Subsequent methods moved from 2D recognition to tooth-centric 3D reconstruction.  \emph{X2Teeth} \cite{x2teeth} explicitly factorizes the problem: an initial stage identifies individual teeth on the PX, and a second stage maps each tooth to a three-dimensional shape drawn from a CBCT-derived template library.  By reconstructing teeth one by one, the method captures detailed morphology and improves over direct volumetric regression.  However, only the dental crowns and roots are modelled; the surrounding bone is not reconstructed, and training requires paired PX--CBCT data for every case.

Related frameworks treat PX as a guidance signal for completing or refining existing 3D scans.  \emph{ToothInpaintor} \cite{ToothInpaintor} starts from a partial 3D dental model with missing teeth and uses the corresponding PX as a constraint, iteratively adjusting the geometry so that the projection of the completed model aligns with the radiograph.  \emph{PX2Tooth} \cite{px2tooth_2024} adopts a similar tooth-wise perspective: each tooth is segmented in the PX and reconstructed as a 3D point cloud using a sizeable collection of paired 2D--3D examples (on the order of hundreds of cases).  While these approaches demonstrate that PX can drive high-quality tooth reconstruction, they remain restricted to individual teeth and do not aim to recover the full dentoalveolar complex, limiting their direct applicability to tasks such as implant planning where bone morphology is critical.

\subsection{Template and Arch-Based Full-Jaw Modelling}

A separate line of work attempts to extend PX-based reconstruction beyond isolated teeth by injecting stronger anatomical priors on jaw shape.  Rather than predicting a full three-dimensional density from scratch, these methods often rely on template geometries, predefined arch curves, or generic jaw models.

In \emph{Oral-3D} \cite{song_2021}, a generative adversarial network (GAN) is first trained to convert a single PX into a flattened volumetric representation of the jaw.  A subsequent deformation step then warps this flattened volume along a dental arch curve extracted from the PX, thereby imposing the natural curvature of the mandible or maxilla.  This two-stage design improves over purely convolutional baselines by enforcing a global jaw shape, but it depends heavily on hand-crafted arch extraction and assumes near-normal anatomy.

\emph{OralViewer} \cite{liang_2021a} pursues a more educational and visualization-oriented objective.  It combines CNN-based estimates of tooth geometry with generic models of gum and jaw to produce a plausible three-dimensional scene suitable for patient communication.  Similarly, \emph{3DentAI} \cite{3dentai} uses an attention U-Net to obtain a flattened jaw volume and then maps this volume onto a curved surface using explicit geometric priors.  While these systems generate visually convincing jaws, their reliance on predefined templates and assumed curvature means that the reconstructed bone is not fully patient specific.  They can struggle in the presence of significant anatomical variation or deformity, and the flatten–warp pipeline may introduce distortions that are difficult to control.

\subsection{Implicit Volumetric Fields and X-Ray Physics}

More recent approaches shift focus from explicit templates towards continuous volumetric representations that are constrained directly by the X-ray formation process.  The central idea is to represent the jaw as an implicit attenuation field $\sigma(\mathbf{x})$ and to enforce that the rendered PX intensities are consistent with the Beer--Lambert law.  Methods such as \emph{NeBLa} \cite{park_2024} and \emph{Oral-3Dv2} \cite{song_2023} embed this physical model into the training objective, thereby reducing reliance on voxel-wise annotated 3D labels and eliminating hand-crafted arch curves \cite{Ketcham_2014,Max_1995}.

Given a ray $\mathbf{r}_{i,j}$ corresponding to pixel $(i,j)$, the predicted PX intensity is modelled as
\begin{align}
	I(i,j) \;=\; I_{0} \exp\Bigl(- \textstyle\int_{t_{\min}}^{t_{\max}} \sigma\bigl(\mathbf{r}_{i,j}(t)\bigr)\, dt\Bigr),
	\label{eq:beer_law_related}
\end{align}
where $I_{0}$ denotes the incident flux.  During training, synthetic PX images are rendered from CBCT volumes using Eq.~\eqref{eq:beer_law_related}, and the parameters of the implicit field are optimized so that these line integrals reproduce observed PX intensities.  In practice, this yields state-of-the-art reconstructions while avoiding the need to specify an explicit dental arch \cite{park_2024,song_2023}.

A key complication arises because panoramic rays are not independent: multiple rays typically traverse the same spatial location, potentially implying different local attenuation values.  \emph{NeBLa} resolves this by introducing an intermediate, ray-aggregated density,
\begin{align}
	\rho(I,\mathbf{x})
	\;=\;\frac{1}{\lvert B(\mathbf{x})\rvert}
	\sum_{\mathbf{p}\in B(\mathbf{x})}
	F\bigl(\mathbf{x},e(I,\mathbf{p})\bigr),
	\label{intermediate_den}
\end{align}
where $B(\mathbf{x})$ is the set of PX pixels whose associated rays intersect $\mathbf{x}$, and $F(\mathbf{x},e(I,\mathbf{p}))$ is a neural prediction conditioned on the image $I$ and a positional encoding $e(I,\mathbf{p})$.  Averaging over $B(\mathbf{x})$ reduces inconsistencies between rays but also smooths high-frequency structure and requires that all intersecting rays be processed and stored, which increases both memory use and computation.

\emph{ViT-NeBLa} modifies both the sampling strategy and the feature-extraction backbone to address these limitations \cite{vit_nebla_2025}.  Instead of using a generic fan-beam, it traces rays along a patient-specific elliptical trajectory that closely follows the dental arch and restricts sampling to a horseshoe-shaped focal region.  The geometry is chosen so that rays do not intersect within the region of interest, removing the need for the intermediate aggregation in Eq.~\eqref{intermediate_den}.  Empirically, this design sharpens tooth and bone boundaries and reduces the number of samples per ray by roughly 52\,\%, leading to more efficient volumetric rendering.  A Vision Transformer (ViT) encoder provides globally consistent features from the PX, which are decoded by a NeRF-like implicit field to obtain the attenuation map.  The main drawback of \emph{ViT-NeBLa}, however, is its training cost: as in other NeRF-style architectures, the model must iteratively estimate attenuation at a large number of sampled 3D points using repeated rendering passes, resulting in substantially longer training times than purely CNN-based PX-to-3D approaches.

\subsection{Architectural Trends: Beyond Pure CNNs}

Across these lines of work, architectural choices have gradually evolved from conventional CNNs to more expressive hybrids and transformer-based models.  Hybrid networks such as \emph{3DPX} \cite{3dpx_2024} combine multi-layer perceptrons with convolutional backbones in a multi-scale framework, allowing information to flow between fine and coarse resolutions and improving depth consistency in the reconstructed volume.  Although these models alleviate some of the limitations of purely local convolutional processing, they still inherit CNNs' difficulty in capturing long-range dependencies from a single, globally aggregated PX.

Self-attention mechanisms, instantiated in Vision Transformers, have therefore attracted increasing interest for PX-guided reconstruction.  ViT-based encoders have already been explored in general 3D reconstruction and radiance-field learning \cite{vit_nerf}, and ViT-NeBLa \cite{vit_nebla_2025} represents a first attempt to pair transformer-style global context modelling with a physics-based implicit field in the panoramic setting.  In principle, such architectures are well suited to resolving the severe depth ambiguity inherent in PX, as they can simultaneously attend to widely separated regions of the image and integrate these cues when predicting the underlying 3D density.

\subsection{From Existing Limitations to the K–U-KAN Framework}

The methods reviewed above trace a clear evolution in PX–guided reconstruction, but they also expose several structural limitations that motivate our proposed framework.  Tooth–centric approaches, such as \emph{X2Teeth}, \emph{ToothInpaintor} and \emph{PX2Tooth} \cite{x2teeth,ToothInpaintor,px2tooth_2024}, demonstrate that a single PX contains sufficient information to recover detailed tooth morphology.  However, they explicitly restrict their scope to the dentition and do not attempt to reconstruct the surrounding alveolar bone.  This omission is critical: for implant planning, assessment of periapical pathology, or evaluation of cortical integrity, the geometry and density of bone are at least as important as the teeth themselves.  As a result, these methods, while technically impressive, cannot directly serve as comprehensive tools for 3D treatment planning.

Template– and arch–driven techniques, including \emph{Oral-3D}, \emph{OralViewer} and \emph{3DentAI} \cite{song_2021,liang_2021a,3dentai}, attempt to fill this gap by imposing strong priors on jaw shape through predefined arches, generic gum and jaw models, or flatten–and–warp pipelines.  In doing so, they extend reconstruction from isolated teeth to an entire jaw–like structure.  Yet this gain in coverage comes with a trade–off: bone morphology is no longer fully patient–specific, and the reliance on auxiliary inputs (manually extracted arch curves, CBCT flattening, or curated template libraries) makes deployment more cumbersome and brittle, particularly in the presence of anatomical abnormalities or domain shifts between synthetic and clinical PX images.  Many CNN–centric architectures in this class also rely on local receptive fields and thus struggle to capture the global context that is essential for resolving depth ambiguity in complex panoramic projections.

Physics–based implicit representations, exemplified by \emph{NeBLa}, \emph{Oral-3Dv2} and \emph{ViT-NeBLa} \cite{park_2024,song_2023,vit_nebla_2025}, move closer to an ideal formulation by embedding the Beer--Lambert law directly into the learning objective and representing the jaw as a continuous attenuation field.  This reduces dependence on dense 3D supervision and removes the need for hand–crafted arch templates, while achieving state–of–the–art quantitative and qualitative reconstruction quality.  However, these approaches introduce their own set of challenges.  In \emph{NeBLa}, intersecting rays force the model to reconcile conflicting local estimates via an intermediate, ray–averaged density \cite{park_2024}, which can blur fine anatomical structures and increases memory and computational cost.  NeRF–style architectures, including ViT-NeBLa, provide strong predictive capability but incur long training times because the attenuation coefficient at each sampled 3D point is learned iteratively through repeated volumetric rendering.  By contrast, purely CNN–based models train much faster but lack the representational power and physical fidelity of implicit fields, leading to a persistent trade–off between accuracy and efficiency.

Across these families of methods, a common pattern emerges: current PX–to–3D systems tend to (i) focus on teeth while under–representing patient–specific bone, (ii) rely on auxiliary information such as dental arch curves, CBCT flattening procedures, or large paired PX–CBCT datasets that may not be available in routine practice, (iii) either struggle with global anatomical context (in the case of CNN–dominated architectures) or suffer from high computational cost and long training times (in the case of NeRF–like implicit fields), and (iv) depend on additional positional encoding modules (Fourier features, hash grids, etc.) to stabilize learning in 3D.  These limitations highlight a gap for methods that can reconstruct the full dentoalveolar complex from a single PX, without extra inputs, while balancing physical interpretability, global context modelling, and computational practicality.

The K–UKAN framework proposed in this article is designed to address this gap.  It aims to reconstruct both teeth and bone from a single panoramic radiograph without any auxiliary information such as arch curves, flattened CBCT volumes, or large-scale paired supervision.  To this end, K–UKAN combines Koopman operator–based learning with Kolmogorov–Arnold Network (KAN) architectures to model the underlying attenuation dynamics in a compact and expressive manner, and adopts an elliptical, non–intersecting sampling strategy that avoids the need for intermediate density aggregation.  Unlike many implicit neural representation–based models, our approach does not require separate positional-encoding modules (e.g., frequency encodings or learnable hash grids); positional information is handled natively within the KAN-based representation.  Empirically, the resulting model achieves qualitative and quantitative performance comparable to existing state–of–the–art implicit methods, while reducing training and inference time relative to NeRF–style architectures.  In this way, K–UKAN advances single–view dental reconstruction toward a more clinically practical, radiation–efficient tool that delivers patient–specific 3D information from the imaging modality most commonly available in routine dental care.

\section{Background: K-UKAN framework basics} \label{sec:background}

\subsection{Koopman Operator Formalism}


Nonlinear dynamical systems often exhibit rich behaviors that are difficult to analyze or exploit directly in state space, especially when one seeks global prediction or control rather than local linearization \cite{Mezic2005,Brunton2022ModernKoopman}. Koopman theory provides an alternative viewpoint by reformulating the dynamics in terms of observables (measurable functions of the state) whose evolution is governed by a linear operator, even when the underlying system is strongly nonlinear \cite{Koopman1931,vonNeumann1932,Mezic2005,Brunton2022ModernKoopman}. In essence, Koopman’s idea is to lift the original nonlinear state-space dynamics into a (typically higher-dimensional and, in general, infinite-dimensional) observable space in which the evolution becomes linear, enabling spectral decompositions and data-driven linear surrogates for nonlinear dynamics \cite{Koopman1931,Rowley2009,Williams2015EDMD}. This perspective has recently been extended through neural and structure-preserving Koopman learners for complex nonlinear systems and PDEs \cite{Xiong2024KNO,Meng2024InvertibleKoopman,Zhang2024HNKO,Sun2025RecursiveRegulator}.

Mathematically, consider an autonomous continuous-time dynamical system evolving on a finite-dimensional state space:
\begin{equation}
	\dot{x}(t)=f\!\big(x(t)\big), \qquad x(t)\in \mathbb{R}^n .
\end{equation}
The associated flow map $F_t:\mathbb{R}^n\to\mathbb{R}^n$ is defined by the state reached after time $t$,
\begin{equation}
	x(t_0+t)=F_t\!\big(x(t_0)\big).
\end{equation}
In discrete time, the dynamics are given by an iterated map
\begin{equation}
	x_{k+1}=F(x_k),
\end{equation}
where $F$ advances the state by one timestep. Koopman theory shifts attention from the state to observables, i.e., measurable functions of the state,
\begin{equation}
	g:M\to \mathbb{R},
\end{equation}
(and, more generally, vector-valued $g$), and defines the Koopman operator through composition with the evolution. In continuous time,
\begin{equation}
	(K_t g)(x)=g\!\big(F_t(x)\big), \label{eq:koopman1}
\end{equation}
while in discrete time,
\begin{equation}
	(K g)(x_k)=g\!\big(F(x_k)\big).\label{eq:koompan_2}
\end{equation}
Thus, the Koopman operator advances measurements along trajectories---``evolve the state, then measure.'' Crucially, although $F_t$ (or $F$) may be nonlinear in $x$, the induced Koopman operator is linear on the space of observables:
\begin{equation}
	K_t(\alpha g_1+\beta g_2)=\alpha K_t g_1+\beta K_t g_2,
\end{equation}
thereby providing a linear (typically infinite-dimensional) representation of nonlinear dynamics in the lifted observable space.

Koopman-based learning has already been deployed in imaging contexts where each frame or slice is treated as a high-dimensional observable of an underlying process. Deep Koopman embeddings learn a latent observable space with approximately linear evolution from raw visual data, enabling stable long-horizon modeling directly from images or videos \cite{lusch2018deepkoopman,wang2023neural_koopman_pooling}. In medical and volumetric imaging, finite Koopman/DMD approximations have been used to propagate sparse 2D slice information via a learned linear operator, allowing interpolation of missing slices and reconstruction of coherent 3D volumes \cite{jo2022dmd_reconstruction,tirunagari2017wr_dmd_mri}. Relative to generic deep state-space models (SSMs), Koopman architectures impose an explicit linear evolution core in the lifted space, which recent benchmarks show can match or exceed SSM performance in nonlinear long-sequence regimes with fewer parameters and improved spectral interpretability \cite{diaz2024ssm_vs_koopman}.

\subsection{Kolmogorov--Arnold Networks and U-KAN}

Kolmogorov--Arnold Networks (KANs) \cite{wang2025kainn,liu2025kan,liu2024kan2,somvanshi2025survey,bozorgasl2024wavkan,blealtan2024efficientkan,moradzadeh2025ukan,li2025ukan,patra2024pikan,sen2025ehrenfestkan}
are neural networks whose design is
directly inspired by the Kolmogorov--Arnold representation theorem:
instead of placing a fixed nonlinearity on each neuron, KANs move the
nonlinearity onto the \emph{edges} of the network. In a standard dense
layer, a neuron first forms a linear combination of its inputs
\begin{equation}
	z = \sum_{i=1}^{n} w_i x_i + b , \label{eq:dense_layer}
\end{equation}
and then passes this scalar through a fixed activation function such as
ReLU or $\tanh$. In a KAN layer, every incoming edge carries its own
learnable univariate function $\phi_i(\cdot)$, typically implemented as
a compact spline or other basis expansion. The neuron simply sums these
transformed inputs,
\begin{equation}
	z = \sum_{i=1}^{n} \phi_i(x_i) + b ,
\end{equation}
so the nonlinearity is already embedded in the edge functions and there
is no need for a separate activation. This edge–centric view is
equivalent to viewing the network as a learnable superposition of
one–dimensional functions, closely mirroring the structure guaranteed by
Kolmogorov--Arnold-type universal approximation theorems.

Stacking such layers yields a deep KAN, where each layer is described by
a function matrix $\phi^{(\ell)}$ that maps an input vector to an output
vector via these sums of univariate functions. Composing the $L$ layers
\begin{equation}
	\mathrm{KAN}(x)
	= \bigl(\phi^{(L-1)} \circ \phi^{(L-2)} \circ \cdots \circ \phi^{(0)}\bigr)(x)
\end{equation}
gives the overall network, exactly as in a multilayer perceptron, but
now every ``weight'' is itself a smooth, trainable curve rather than a
scalar. This seemingly small change has two important consequences:
(i)~KANs often reach a desired accuracy with fewer parameters than
standard MLPs and exhibit favorable scaling on data-fitting and PDE
benchmarks; and (ii)~each learned edge-function can be plotted,
inspected and even approximated symbolically, giving KANs a degree of
interpretability that is hard to obtain from ordinary dense layers.
Variants such as Wav-KAN replace splines with wavelet bases to better
capture multiscale structure, while practical libraries package these
ideas into implementations for scientific machine learning.

Building on these foundations, the U-KAN architecture shows how KAN
layers can serve as a strong, interpretable backbone inside
U-Net--style models for medical imaging and diffusion-based generative
models \cite{li2025ukan, zhang2025ushapekanbridge}. The overall encoder–decoder with skip connections is kept from
the original U-Net, but near the bottleneck a \emph{tokenized KAN
	block} is introduced: high-level feature maps are reshaped into tokens,
passed through KAN operators that model rich nonlinear interactions
across channels, and then reshaped back before decoding. This design
directly targets two weaknesses of conventional U-Nets: a limited
ability to capture complex cross-channel nonlinear patterns and poor
interpretability of what each channel represents. On a range of medical
image segmentation benchmarks, U-KAN achieves improved performance over
strong U-Net variants with comparable or even lower computational cost,
and the same KAN block can act as an enhanced noise-prediction backbone
in diffusion models. In this way, U-KAN demonstrates how mathematically
grounded KAN operators can be dropped into standard vision pipelines to
gain accuracy, efficiency, and a more transparent view of what the
network has learned.

\section{Methodology}\label{sec:method}

A panoramic radiograph (PX) packs 3D x-ray attenuation into a 2D image by integrating along curved scanner rays. We invert this packing in three stages: (i) a \emph{learned lift} that turns PX pixels into \emph{observables} (i.e., functions of the hidden anatomy we can compute from features) using Kolmogorov–Arnold Networks (KAN), (ii) a \emph{Koopman step} that advances those observables by a tiny, stable, \emph{linear} evolution—“shrink \& twist with a bounded nudge,” and (iii) an \emph{inverse warp} that places predicted depth bins back into the horseshoe-shaped \emph{focal trough} (the scanner’s in-focus band following the dental arch) to form a volumetric seed for refinement. Each piece mirrors either the physics of image formation (Beer–Lambert), the geometry of the arch (focal trough), or a stability prior (Koopman spectrum).

\subsection{Problem formulation}

Let $\Omega\subset\mathbb{R}^3$ denote the oral volume and $\mu:\Omega\to\mathbb{R}_{\ge 0}$ its linear attenuation field (x-ray absorption per unit length). A PX image $I:\mathcal{U}\subset\mathbb{R}^2\to\mathbb{R}_{\ge 0}$ is generated by Beer–Lambert integration (also defined in Eq. \eqref{eq:beer_law_related}) of $\mu$ along scanner rays $\gamma_u:[0,L_u]\to\Omega$ associated with detector coordinate $u\in\mathcal{U}$:

\begin{align}
	I(u) \;\propto\; \exp\!\left(-\int_0^{L_u}\mu(\gamma_u(s))\,ds\right),
	\qquad u\in\mathcal{U}.
	\label{eq:beer}
\end{align}

In other word, for each detector location $u$, we traverse the ray $\gamma_u$, accumulate attenuation $\mu$ (the integral), then exponentiate to obtain intensity $I(u)$; depth is therefore hidden inside line integrals.

Recovering $\mu$ from single PX image $I \in \mathbb{R}^{1 \times H \times W}$ is ill-posed. We thus factor the reconstruction into three stages which constitutes the whole pipeline of the purposed framework (Vide Fig. \ref{fig:ukan_fw}). (i) lift $I$ to a structured depth-binned representation $F$ over the image grid (Fig. \ref{fig:ukan_fw}(A)), (ii) inversely warp $F$ onto a horseshoe (focal-trough) coordinate chart aligned with the dental arch to obtain a CBCT seed $V^{(0)}$ (Fig. \ref{fig:ukan_fw}(B)), and (iii) refine $V^{(0)}$ with a dedicated 3D module detailed in section \ref{sec:refinement} (Fig. \ref{fig:ukan_fw}(C)). Concretely,

\begin{align}
	\underbrace{I}_{\text{PX}} \;\xrightarrow{\ \mathcal{F}_\theta\ }\;
	\underbrace{F}_{\text{depth bins}}\;\xrightarrow{\ \Phi^{-1}\ }\;
	\underbrace{V^{(0)}}_{\text{horse-shoe seed}} \; \xrightarrow{\ \mathcal{R}_\phi\!\big(\tilde V^{(0)}\big)\ }\; V,
	\label{eq:pipeline}
\end{align}

where, $\mathcal{F}_\theta$ predicts $K$ \emph{depth bins per pixel}—a ``flat'' depth field $F \in \mathbb{R}^{K\times H\times W}$. The inverse warp $\Phi^{-1}$ moves bins onto precomputed rays that stay inside the focal trough (horseshoe band where PX is sharp), producing a seed $V^{(0)} \in \mathbb{R}^{H\times W\times D}$. The binary mask $M \in \{0,1\}^{H\times W\times D}$ restricts predictions to the focal trough, yielding $\tilde V^{(0)}=M\odot V^{(0)}$ prior to refinement. $\mathcal{R}_\phi$ is a 3D refinement module which refine the predicted seed volume $V^{(0)}$ and finally give the final 3D volume $V \in \mathbb{R}^{H \times W \times  D}$. We use $H{=}128$, $W{=}256$, $K{=}96$, $D{=}256$.

\subsection{Proposed Architecture (Koopman-based U-KAN)}

\begin{figure*}[ht]
	\centering
	\includegraphics[width=\linewidth]{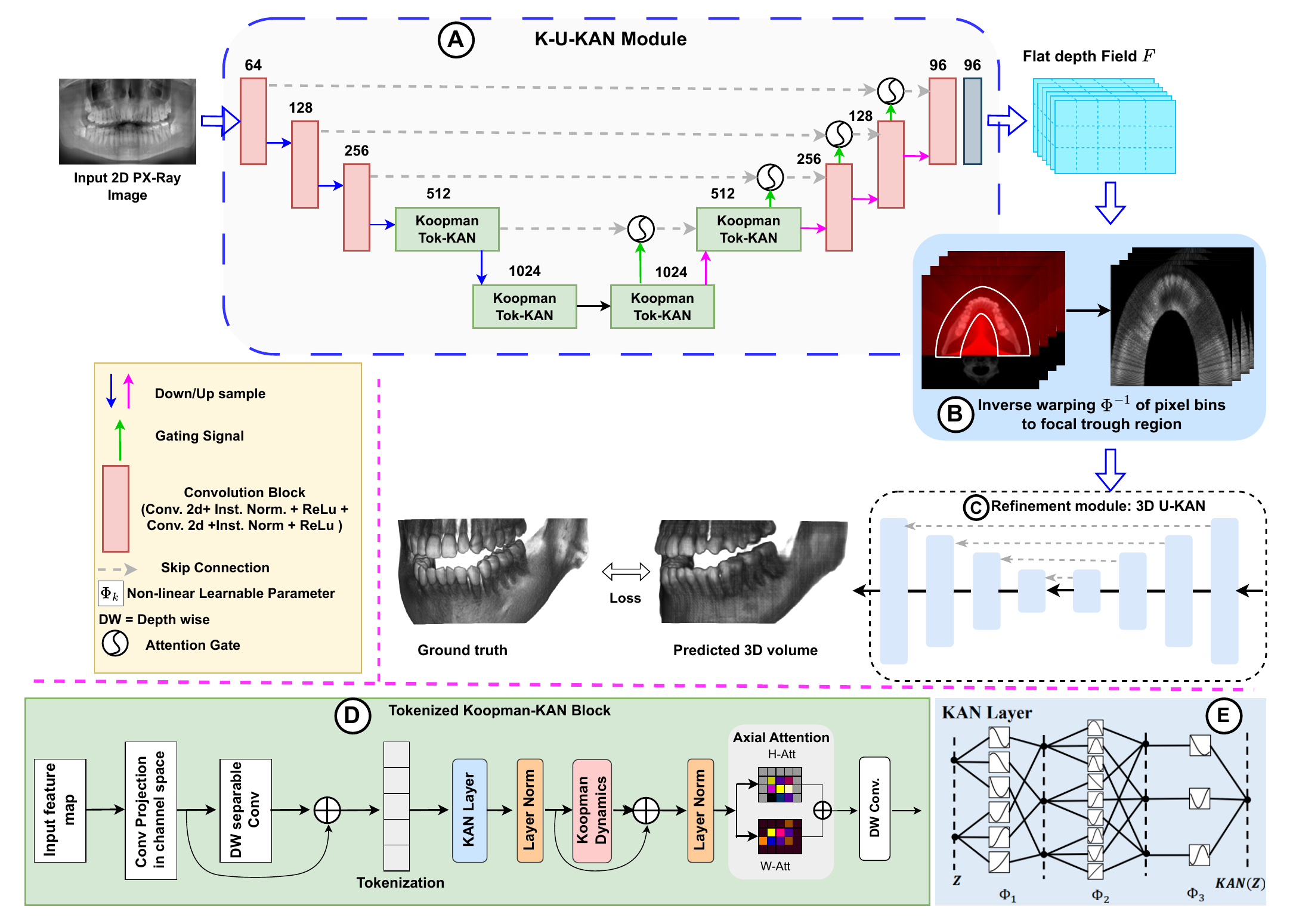}
	\caption{\textbf{Schematic overview of the proposed K-U-KAN framework for single-view 3D dental reconstruction.}
		\textbf{(A) K-U-KAN Lifting Module:} The input 2D panoramic radiograph is processed by a U-Net-style encoder-decoder where bottleneck features are lifted into a latent observable space. The network utilizes Tokenized Koopman-KAN blocks to predict a flat depth field $F$ consisting of $K$ depth bins.
		\textbf{(B) Focal-Trough Inverse Warping:} A geometric transformation $\Phi^{-1}$ maps the predicted depth bins onto precomputed ray trajectories within the dental arch's horseshoe-shaped focal trough, generating an initial volumetric seed $V^{(0)}$.
		\textbf{(C) 3D Attention U-KAN Refinement:} The coarse seed $V^{(0)}$ is processed by a 3D U-KAN refiner ($\mathcal{R}_{\phi}$) equipped with attention gates to recover fine anatomical details and produce the final CBCT volume.
		\textbf{(D) Tokenized Koopman-KAN Block:} Detail of the latent dynamics block, which integrates KAN-based feature lifting and a stable, phase-aware Koopman operator for linear feature evolution.
		\textbf{(E) KAN Layer:} Illustration of the Kolmogorov-Arnold Network layer structure, employing learnable univariate spline functions on edges to model non-linearities.}
	\label{fig:ukan_fw}
\end{figure*}

The lifting network $\mathcal{F}_\theta$ is a \emph{Koopman-aware U-KAN (K--U--KAN)} (illustrated in Fig. \ref{fig:ukan_fw}(A)) \cite{li2025ukan}: a U-Net encoder/decoder with attention-gated skips whose bottlenecks are replaced by \emph{token blocks} that run a KAN lift, a Koopman step, and axial attention.  The workflow of the token block is depicted in the Fig. \ref{fig:ukan_fw}(D).

First, a small conv puts features in a good channel space; depthwise separable conv adds cheap locality; KAN chooses a coordinate system where the next move is simple; Koopman makes that move linearly and safely; axial attention shares information across the long arch; a depthwise conv plus GroupNorm stabilizes batch size $=1$. 

The encoder uses two $3\times3$ convs per block (InstanceNorm, ReLU), downsampling by stride-2 $3\times3$ convs. The decoder mirrors this with transposed $3\times3$ (stride 2); skip connections pass through attention gates. The head maps to $K$ bins, followed by a scaled sigmoid to preserve native radiographic range $[0,255]$.\\

\paragraph{\textbf{Token Lift via Kolmogorov--Arnold Networks (KAN).}}  Given per-pixel features $z\in\mathbb{R}^{D}$ (after the projection conv), a shallow KAN returns observables $\phi(z) = \mathrm{KAN} (\mathbf{z}) \in\mathbb{R}^{D}$ by summing univariate spline responses,

\begin{align}
	\varphi^{(\ell)}_j(x) \;=\; \sum_i \phi^{(\ell)}_{ji}(x_i) + b^{(\ell)}_j,
	\quad
	\phi \;=\; \varphi^{(L-1)}\circ\cdots\circ\varphi^{(0)}(z).
	\label{eq:kan_unit}
\end{align}

Instead of a fixed nonlinearity per neuron as in the standard dense layer defined in Eq. \eqref{eq:dense_layer}, each incoming edge has its own learnable 1-D function (a spline). This lets KAN \emph{re-express} features in coordinates where the next step can be made linear. A visual illustartion of the KAN layer is illustrated in the Fig. \ref{fig:ukan_fw}(E). 

\paragraph{\textbf{Koopman token dynamics.}} At each pixel, we already have a vector of observables $\boldsymbol{\phi}=\mathrm{KAN}(\mathbf{z})$ from Eq.~\eqref{eq:kan_unit}.  We now need a one-step rule that (i) moves $\boldsymbol{\phi}$ forward in a way that is easy to control and reason about, (ii) keeps energy from blowing up on raw $[0,255]$ inputs and batch size $=1$, and (iii) captures gentle, arch-following variations across columns. 
Koopman theory says we can do this by evolving \emph{measurements} linearly: “evolve the state, then measure.”  We do the practical version: \emph{evolve the measurements directly} by a tiny, stable, per-channel linear map.

Imagine each channel $d$ of $\boldsymbol{\phi}$ as a little arrow on the plane.  One step takes that arrow and (a) \emph{shrinks} it a bit so things remain stable, (b) \emph{twists} it a bit to encode phase/curvature along the arch, and (c) applies a \emph{tiny bounded nudge} so we can match real data without ever running away. Then we tidy up with a short linear mix and a normalization to keep scales honest.  That’s all the Koopman block does.

Classical Koopman evolves measurements linearly as $(Kg)(x)=g(F(x))$ (also defined in Eq. \eqref{eq:koopman1} and Eq.\eqref{eq:koompan_2}). If a measurement is special (an eigen-observable), then $g(F(x))=\lambda g(x)$ with $\lambda=|\lambda|e^{i\theta}$; geometrically, \emph{shrink by} $|\lambda|$, \emph{rotate by} $\theta$.

We do exactly that per channel, with a safety brake and a gentle correction:
\begin{align}
	\boldsymbol{\phi}^{+}&=\Lambda\,\boldsymbol{\phi},\qquad
	\Lambda=\mathrm{diag}(\lambda_1,\ldots,\lambda_D),
	\label{eq:diag}\\
	|\lambda_d|&=\exp\!\big(-\mathrm{softplus}(\nu_d)\big)\in(0,1),\quad
	\angle\lambda_d=\theta_d\in\mathbb{R},
	\label{eq:lambda}
\end{align}
so magnitudes are \emph{strictly} below one (contractive) and phases are learned. Because channels are only approximately eigen-observables, we add a bounded, content-adaptive gate computed from the damped magnitude $\mathbf{m}=\log(1+|\boldsymbol{\phi}|)$:

\begin{align}
	g_{r,d}&=1+\rho\,\tanh(\mathbf{a}_{r,d}^{\!\top}\mathbf{m}+b_{r,d}), \nonumber\\
	g_{i,d}&=\rho\,\tanh(\mathbf{a}_{i,d}^{\!\top}\mathbf{m}+b_{i,d}),\quad 0<\rho<1,
	\label{eq:koop_gate}
\end{align}

so $g_d=g_{r,d}+i g_{i,d}$ is near $1{+}0i$ and uniformly bounded. The total complex multiplier is $\alpha_d=\lambda_d g_d$.

Since features are real, the Koopman block realizes this complex scaling by two real elementwise products:
\begin{equation}
	\mathbf{u}=\Re(\boldsymbol{\alpha})\odot\boldsymbol{\phi},\qquad
	\mathbf{v}=\Im(\boldsymbol{\alpha})\odot\boldsymbol{\phi},
	\label{eq:split}
\end{equation}
stack $[\mathbf{u};\mathbf{v}]\in\mathbb{R}^{2D}$, apply a small linear projection $\mathbf{W}\in\mathbb{R}^{D\times 2D}$, add a residual, and normalize:
\begin{equation}
	\boldsymbol{\phi}^{+}=\mathrm{LN}\!\big(\boldsymbol{\phi}+\mathbf{W}\,[\mathbf{u};\mathbf{v}]\big).
	\label{eq:koop_residual}
\end{equation}

If we turn off the nudge ($\rho{=}0$) and let $\mathbf{W}$ pick the needed component; Eq.~\eqref{eq:koop_residual} reduces to $\boldsymbol{\phi}^{+}=\Lambda\boldsymbol{\phi}$—the eigen-observable update i.e. multiply each measurement by its eigen-value (Koopman's $g(F(x)) = \lambda g(x)$ ). With the nudge on, we stay \emph{near} that ideal but better fit data because $|g_d-1|\le\rho$. Since $|\lambda_d|<1$ and $g_d$ is bounded, $|\alpha_d|$ remains close to contractive, which prevents energy build-up across many steps and $[0,255]$ activations.

The learned phases $\theta_d$ act like tiny per-column rotations (a phase velocity). As we sweep across the detector, these micro-rotations let channels \emph{track the curvature} of the dental arch. After the inverse warp $\Phi^{-1}$, this manifests as \emph{depth-consistent placement} along focal-trough rays.

\paragraph{\textbf{Axial Self-Attention for Long-Range Context.}} To capture long-range arch geometry without quadratic complexity in $HW$, we apply multi-head self-attention independently along width and height and sum the outputs:
\begin{equation}
	\mathrm{AxialAttn}(X)=\mathrm{MHA}_w(X)+\mathrm{MHA}_h(X),
	\label{eq:axial}
\end{equation}

with per-axis LayerNorm. This global context complements local depthwise operations within each token block and empirically improves arch-consistent depth allocation. Complexity is $O(HWC(H{+}W))$, not $O((HW)^2)$.

\subsection{Inverse warping to the focal trough}

We parametrize the focal trough in the axial plane by confocal ellipses with center $(h,k)$ and semi-axes $(a_1,b_1)$ (outer) and $(a_2,b_2)$ (inner). A point $(x,y)$ lies inside the outer trough if
\begin{equation}
	\frac{(x-h)^2}{b_1^2}+\frac{(y-k)^2}{a_1^2}\le 1.
	\label{eq:ellipse}
\end{equation}
For each image column $i\in\{0,\ldots,W-1\}$, we precompute a ray $\Gamma_i(t)=\mathbf{o}_i+t\,\mathbf{d}_i$ and retain the first $K$ intersections $\{(x_{i,k},y_{i,k})\}_{k=1}^{K}$ satisfying Eq. \eqref{eq:ellipse}. The flat field $F(k,j,i)$ (depth bin $k$, row $j$, column $i$) is \emph{scattered} to axial voxels:

\begin{equation}
	V^{(0)}(y_{i,k},x_{i,k},j)\ \xleftarrow{\Phi^{-1}}\ F(k,j,i),
	\quad \forall\,i,j,k,
	\label{eq:scatter}
\end{equation}

then permuted to $[1\times H\times W\times D]$ and masked: $\tilde V^{(0)}=M\odot V^{(0)}$. Precomputing $\{(x_{i,k},y_{i,k})\}$ amortizes geometry across the batch and reduces training time. The overall pipeline of this module is illustrated in the Fig. \ref{fig:ukan_fw}(B) and Fig. \ref{fig:wf_data_pre}(C).

\subsection{Refinement Module : Attention 3D U-KAN} \label{sec:refinement}


The volumetric refiner $\mathcal{R}_\phi$ upgrades the masked seed $\tilde V^{(0)}\!\in\!\mathbb{R}^{1\times H\times W\times D}$ into a high-fidelity CBCT $V\!\in\!\mathbb{R}^{1\times H\times W\times D}$ using a 3D U\textendash Net scaffold augmented with a tokenized KAN bottleneck and attention-gated skips \cite{li2025ukan,zhou_u-net_2020,att_u_net}. The encoder comprises four stages, each applying two $3{\times}3{\times}3$ convolutions with InstanceNorm3d and Swish activations (robust for batch size $=1$), followed by a stride-$2$ $3{\times}3{\times}3$ downsampling convolution. Channel widths are $(32,\,64,\,128,\,256)$, and the deepest stage maps to $d{=}512$ channels.

At the coarsest scale ($H/16\times W/16\times D/16$; e.g., $8{\times}16{\times}16$ for $128{\times}256{\times}256$ volumes), features are projected by a $3{\times}3{\times}3$ convolution and reshaped into tokens for a \emph{Tok\textendash KAN 3D} block. Concretely, a shallow KAN with hidden width $D\!=\!d$ (grid size $5$, spline order $3$, SiLU bases) produces per-token observables; a depthwise 3D convolution (groups$=D$) injects local spatial coupling; and a residual LayerNorm in token space stabilizes mixed precision. An optional second Tok\textendash KAN block further strengthens long-range channel mixing with minimal parameter growth.

The decoder mirrors the encoder with $3{\times}3{\times}3$ transposed convolutions (stride $2$) for upsampling. Each level concatenates the corresponding encoder features filtered through 3D attention gates (additive gating of skip tensors conditioned on the current decoder context) before two $3{\times}3{\times}3$ Conv\textendash InstanceNorm3d\textendash Swish layers. A final $3{\times}3{\times}3$ head reduces to a single output channel, and a calibrated logistic $x\mapsto 255\,\sigma(0.5x)$ preserves the native radiographic range.

Practically, this design (i) retains the spatial precision and multi-scale context of a U\textendash Net, (ii) concentrates KAN’s edge-wise spline nonlinearities where capacity is most valuable (the bottleneck), and (iii) uses attention gates to pass only task-relevant encoder detail. In combination, these choices yield sharper enamel\textendash dentin boundaries and more coherent mandibular canals from the coarse seed.

\subsection{Loss functions}

To balance local numerical accuracy with anatomically plausible global structure and visually realistic projections, we train the network using a composite objective that combines voxel-wise, projection-based, and perceptual criteria.


The network parameters are optimized using a composite objective that balances voxel-level fidelity, global structural consistency, and perceptual realism.  Concretely, for an input PX image $I$ and its corresponding CBCT volume, we employ three complementary loss terms: a volumetric mean squared error, a multi-view projection loss, and a feature-based perceptual loss.  The overall objective is

\begin{align}
	\mathcal{L} \;=\; \mathcal{L}_{MSE} \;+\; \lambda_{1}\,\mathcal{L}_{proj} \;+\; \lambda_{2}\,\mathcal{L}_{perc}\,,
	\label{tot_loss}
\end{align}

where $\lambda_{1},\lambda_{2} \ge 0$ control the relative strength of the projection and perceptual components, respectively.  Each term is described below.

\paragraph{\textbf{Volumetric MSE Term ($\mathcal{L}_{MSE}$).}}
The first term enforces accurate reconstruction at the level of individual voxels.  Let $\rho(I)$ denote the predicted density for input $I$, and let $\hat{\rho}(I)$ be the corresponding ground-truth density.  Over the voxel grid $\Omega$ of size $|\Omega|$, we define

\begin{align}
	\mathcal{L}_{MSE} \;=\; \frac{1}{|\Omega|}\sum \Bigl(\rho(I) \;-\; \hat{\rho}(I)\Bigr)^2\,.
\end{align}

This term penalizes local discrepancies between the predicted and reference volumes and ensures that the reconstructed attenuation values are numerically close to the ground truth at every voxel.

\paragraph{\textbf{Multi-View Projection Term ($\mathcal{L}_{proj}$).}}
While the volumetric MSE operates in 3D, it does not explicitly constrain how the reconstruction appears in standard anatomical views.  To promote global consistency, we compare maximum-intensity projections (MIPs) of the predicted and ground-truth volumes along the axial, sagittal, and coronal directions.  For a volume $V$ and a view $v \in \{\text{axial},\,\text{sagittal},\,\text{coronal}\}$, the MIP is denoted by $\Pi_v(V)$; for example, the axial projection is

\begin{align}
	\Pi_{\text{axial}}(V)(x,y)=\max_{z}\; V(x,y,z),
\end{align}

with analogous definitions for the sagittal and coronal views.  The projection consistency loss is then

\begin{align}
	\mathcal{L}_{proj} \;=\; \sum_{v \in \left\{ \substack{\text{axial},\\[3pt] \text{sagittal},\\[3pt] \text{coronal}} \right\}} \;\bigl\|\,\Pi_v\bigl(\rho(I)\bigr) \;-\; \Pi_v\bigl(\hat{\rho}(I)\bigr)\bigr\|_{2}^{2}\,,
\end{align}

where $\|\cdot\|_{2}^{2}$ denotes the sum of squared differences over all pixels in the projection image.  By matching these MIPs, we encourage the predicted volume to reproduce the global anatomical layout observed when the data are viewed in common clinical planes.

\paragraph{\textbf{Perceptual Feature Term ($\mathcal{L}_{perc}$).}}
Finally, pixel-level agreement in the projection domain does not fully capture perceptual similarity or subtle textural patterns.  To address this, we introduce a perceptual loss defined in the feature space of a pretrained VGG-16 network $\Theta$.  Let $\Theta_{\ell}(I)$ denote the activation at layer $\ell$ for an input image $I$, and let $L$ be a selected set of layers.  For each view $v$, we compute feature maps for the MIPs of the predicted and ground-truth volumes, $\Pi_v\bigl(\rho(I)\bigr)$ and $\Pi_v\bigl(\hat{\rho}(I)\bigr)$, and define

\begin{align}
	\mathcal{L}_{perc} 
	&\;=\;\nonumber\\
	&\sum_{v \in \left\{ \substack{\text{axial},\\[3pt] \text{sagittal},\\[3pt] \text{coronal}} \right\}} 
	\sum_{\ell \in L} 
	\Bigl\| \,\Theta_{\ell}\bigl(\Pi_v\bigl(\rho(I)\bigr)\bigr) \;-\; 
	\Theta_{\ell}\bigl(\Pi_v\bigl(\hat{\rho}(I)\bigr)\bigr)\Bigr\|_{2}^{2}\,.
\end{align}

This term measures discrepancies between predicted and reference projections in a deep feature space, encouraging the model to preserve higher-level structures and fine anatomical details that are important for perceptual quality but may not be fully captured by voxel-wise or pixel-wise losses alone.

\section{Experiments} \label{sec:expt}

\subsection{Dataset}
\paragraph{\textbf{Dataset collection.} }

The clinical dataset used in this study consists of 623 cone-beam CT examinations of the head region acquired at the Chosun School of Dentistry in Gwangju, South Korea.  All data were collected under institutional review board approval with written informed consent.  The cohort includes 408 female and 215 male subjects, with a mean age of $50 \pm 26$ years (range: 16–99 years).  

Imaging was performed on two CBCT systems: the Carestream CS9000, which contributed the majority of cases (608 volumes), and the Planmeca VISOG7, which accounted for the remaining 15 scans.  All studies were exported as 16-bit DICOM series.  The reconstructed volumes exhibit heterogeneous field-of-view sizes, with matrix dimensions ranging from $337 \times 337 \times 214$ to $673 \times 673 \times 568$ voxels.  In-plane pixel spacing varied between 0.25~mm and 0.50~mm, while the slice increment was 0.25~mm for 96\% of scans, 0.30~mm for 3.5\%, and between 0.45~mm and 0.50~mm for fewer than 0.5\% of cases.  


\paragraph{\textbf{Dataset preparation.}}

\begin{figure*}[ht]
	\centering
	\includegraphics[width=\linewidth]{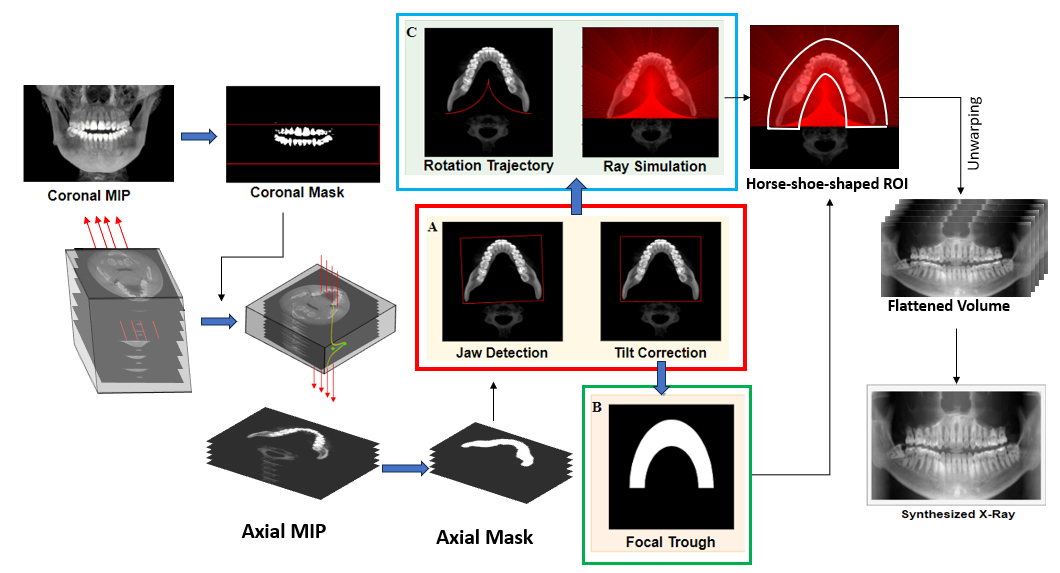}
	\caption{Overview of the data-preparation workflow \cite{Anu_2024,vit_nebla_2025}. Coronal and axial MIPs are first generated to localize the jaw. A. Jaw contour segmentation followed by horizontal tilt correction to standardize alignment. B. Definition of an elliptical focal-trough region that encompasses the dental arch. C. Computation of a dynamic rotation trajectory and simulation of pencil-beam X-ray projections. D. Synthesis of the final panoramic X-ray image from the simulated projections.}
	\label{fig:wf_data_pre}
\end{figure*}

The overall data processing workflow is illustrated in Fig.~\ref{fig:wf_data_pre}.  Following the recent work \cite{Anu_2024,vit_nebla_2025}, we first resample all clinical CBCT volumes to an isotropic voxel spacing of $0.3\,\text{mm}$.  An automatic preprocessing step is then applied to each resampled scan to delineate a contour-based bounding box that tightly encloses the jaw region.  This bounding box defines the three-dimensional region of interest (ROI) used for subsequent reconstruction.  The ROI is cropped from the full CBCT and resized to a fixed grid of $128 \times 256 \times 256$ ($H \times W \times D$), providing a standardized input geometry across patients.

To obtain the corresponding panoramic inputs, we simulate PX images using a forward projection model similar in spirit to the panoramic geometry described by Park et al.\ \cite{park_2024}.  A panoramic trajectory is defined and rays are traced through the resampled ROI to synthesize PX images of size $128 \times 256$.  In contrast to the circular or generic fan-beam trajectories used in \cite{park_2024}, we adopt an elliptical orbit whose rays are tangent to the trajectory and restricted to a horseshoe-shaped focal region that tightly follows the dental arch.  Only samples within this focal region are retained, ensuring that the ray sampling is concentrated on the jaw and avoiding unnecessary passes through irrelevant anatomy.

The attenuation samples collected along each ray within the focal region are then reparameterized on a regular lattice to form a flattened jaw volume of size $128 \times 256 \times 96$ ($H \times W \times D$).  This canonical shape is chosen to match the volumetric output resolution of the proposed K--U-KAN network.  After prediction, the flattened representation is inversely warped back into the original horseshoe-shaped coordinate system, thereby restoring the natural jaw curvature, and the resulting volume is passed to a refinement module for further enhancement.  By defining this step we eliminated the need for use of the flatttened volume data and directly trained our model on cropped CBCT region of interest in the anatomically meaningful horseshoe-shaped space. 

Model training thus exploits the paired representation of each case: the standardized ROI CBCT volume and its synthesized PX image.  Of the 623 available CBCT scans, 600 volumes satisfied the preprocessing and coverage criteria and were successfully converted into CBCT–PX pairs using the above procedure.  These 600 cases were partitioned into training, validation, and test sets using an $8{:}1{:}1$ split.

\paragraph{\textbf{Data pre-processing.}}

In volumetric medical imaging, and particularly in the setting of 3D CBCT volumes and their associated 2D PX images, the raw intensity values are often not directly suitable for learning-based methods.  Differences in scanner protocols, reconstruction settings, and patient anatomy can lead to highly variable histograms with heavy tails and outliers.  To reduce this variability and improve numerical conditioning, we apply a simple but effective sequence of intensity transformations prior to network training.

Let $\mathcal{I}\in\mathbb{R}^{C\times H\times W\times D}$ denote a CBCT volume with a single channel ($C=1$, $H=128$, $W=256$).  As a first step, the voxel intensities are truncated to the range defined by the 1\textsuperscript{st} and 99.9\textsuperscript{th} percentiles of the empirical distribution for that scan, thereby suppressing extreme values that might otherwise dominate the statistics.  The clipped volume is then standardized using Z-score normalization: we subtract the global mean intensity and divide by the global standard deviation, which centers the distribution and enforces unit variance.  In the final stage, the standardized intensities are mapped linearly onto the interval $[0,255]$ so that the resulting values are compatible with common 8-bit image conventions and with pretrained backbones that expect similar ranges.

Overall, this three-step procedure—percentile-based clipping, Z-score standardization, and linear rescaling—reduces the influence of outliers, reshapes the intensity histogram into a more symmetric form, and constrains the dynamic range.  The processed CBCT volumes thus provide numerically stable and tightly normalized inputs for the subsequent 3D reconstruction network.

\subsection{Training protocol}

\begin{table}[htbp!]
	\centering
	\caption{Summary of training configurations.}
	\label{tab:train_config}
	\begin{small}
		\renewcommand{\arraystretch}{1.2}%
		\resizebox{\linewidth}{!}{%
		\begin{tabular}{lc}
			\toprule
			\textbf{Parameter} & \textbf{Value} \\
			\midrule
			Deep Learning Framework  & PyTorch \\
			GPU Configuration       & NVIDIA RTX A6000 \\
			Training Time for 600 samples          & $\sim$10 hours 43 mins \\
			Evaluation Time per sample &  $\sim$ 24 Seconds \\
			Optimizer               & Adam ($\beta_{1}=0.9,\;\beta_{2}=0.999$) \\
			Initial Learning Rate   & $1\times10^{-3}$ \\
			LR Scheduler            & ReduceLROnPlateau \\
			Batch Size              & 1 \\
			Number of Epochs        & 300 \\
			Early Stopping          & Enabled (based on validation loss) \\
			\bottomrule
		\end{tabular}}
	\end{small}
\end{table}

The proposed reconstruction model is implemented in PyTorch and trained on NVIDIA RTX A6000 GPUs.  A full training run requires approximately 10~hours and 43~minutes of wall-clock time, and the average inference time for a single test case is about 24~seconds.  Optimization is performed using the Adam optimizer with an initial learning rate of $1\times10^{-3}$ and momentum parameters $\beta_{1}=0.9$ and $\beta_{2}=0.999$.  To adapt the step size during training, we employ a \texttt{ReduceLROnPlateau} scheduler that monitors the validation loss and halves the learning rate after 15 consecutive epochs without improvement, subject to a minimum learning rate of $1\times10^{-5}$. Unless otherwise stated, all experiments are trained for up to 300~epochs with a batch size of 1, and an early stopping criterion based on validation performance is used to prevent overfitting. A concise overview of the training and optimization settings is provided in Table ~\ref{tab:train_config}.

\subsection{Metrics}
To quantitatively evaluate the quality of the reconstructed volumes and their projections, we employ three widely used full-reference image quality metrics: PSNR, SSIM, and LPIPS.

\begin{itemize}
	\item \textbf{PSNR.} The Peak Signal-to-Noise Ratio (PSNR) quantifies the fidelity of a reconstruction by comparing the maximum possible pixel intensity to the mean squared error between the reconstructed and reference images.  Higher PSNR values indicate that the reconstruction is closer to the ground truth, and the measure is routinely used in the assessment of lossy image reconstruction and compression methods.
	
	\item \textbf{SSIM.} The Structural Similarity Index (SSIM) evaluates image quality by comparing structural information as well as luminance and contrast between two images \cite{SSIM}.  In contrast to purely error-based measures, SSIM is designed to correlate more strongly with human visual perception by emphasizing local pattern consistency rather than absolute intensity differences. 
	
	\item \textbf{LPIPS.} The Learned Perceptual Image Patch Similarity (LPIPS) metric \cite{zhang_2018} measures perceptual discrepancy between a reconstructed image and its reference by comparing deep feature representations rather than raw pixels.  In our experiments, LPIPS scores are computed using feature activations from standard pretrained networks (VGG-16, AlexNet, and SqueezeNet), providing a complementary view of perceptual quality beyond PSNR and SSIM.
\end{itemize}

\subsection{Results and Discussion}

\subsubsection{Qualitative results}

\begin{figure*}[htbp!]
	\centering
	\includegraphics[width=0.99\linewidth]{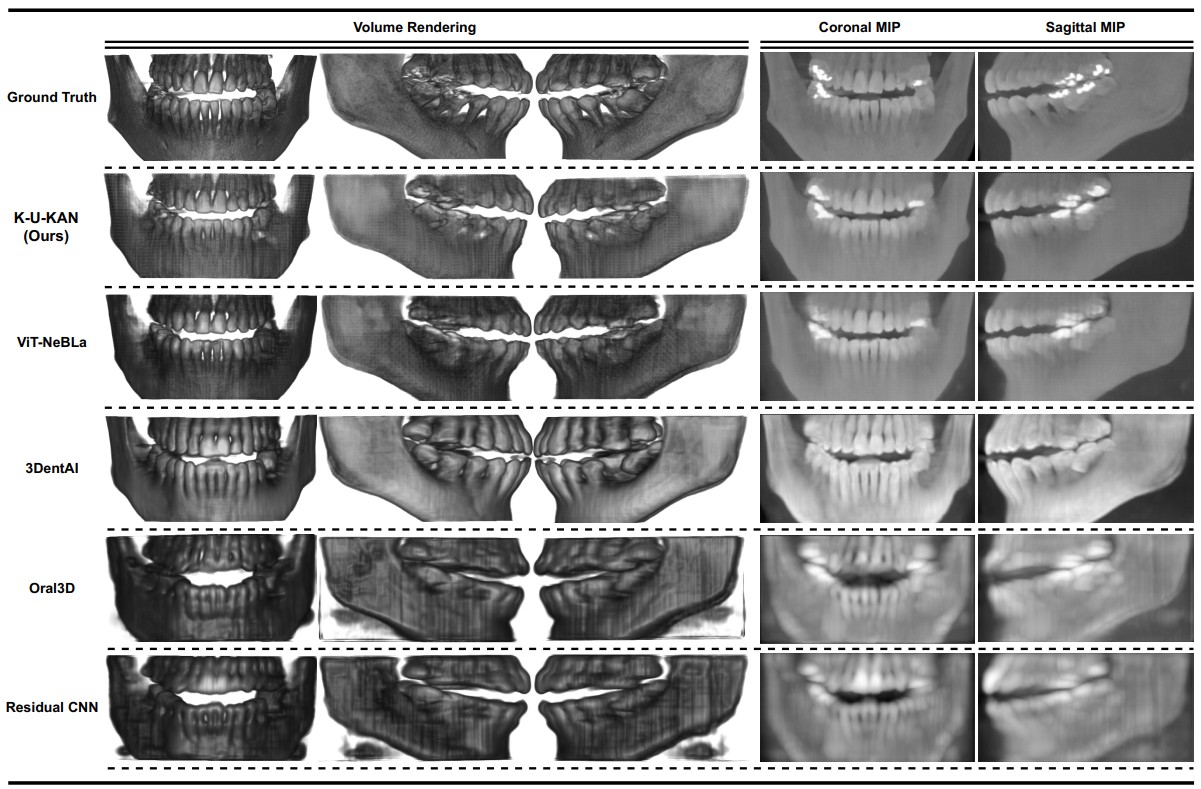}
	\caption{Qualitative comparison of reconstructed CBCT volumes. From top to bottom: ground truth; K-U-KAN (Ours), ViT-NeBLa \cite{vit_nebla_2025}; 3DentAI \cite{3dentai}; Oral-3D (auto-encoder) \cite{song_2021}; residual CNN \cite{henzler_2018}. Columns show (left) volume renderings, (middle) coronal MIPs, and (right) sagittal MIPs. Our K-U-KAN model yields the most accurate delineation of jaw anatomy—preserving cortical boundaries and fine internal structures-while suppressing artifacts and background noise compared to existing methods.}
	\label{fig:qc1}
\end{figure*}

Figure~\ref{fig:qc1} (see Appendix \ref{app:additional_fig} for additional cases) compares 3D volume renderings together with coronal and sagittal MIPs across methods. The proposed \textbf{K–U–KAN} most closely matches the ground truth in both global morphology and fine dental detail: cortical outlines along the inferior border and ramus are sharply resolved, cuspal tips and interproximal contacts remain well separated without haloing, and the mandibular canal trajectory is continuous with minimal breaks. By contrast, \textit{ViT–NeBLa} yields anatomically plausible volumes but exhibits faint banding and mild over-smoothing across the occlusal plane in the sagittal MIP \cite{vit_nebla_2025}; \textit{3DentAI} improves edge definition relative to CNN baselines yet shows sporadic intensity streaks near the condylar and retromolar regions \cite{3dentai}; \textit{Oral-3D} and the \textit{Residual CNN} preserve coarse jaw shape but blur enamel–dentin interfaces and attenuate trabecular texture, producing flattened coronal MIPs and porous volume renders \cite{song_2021,henzler_2018}. Notably, \textbf{K–U–KAN} suppresses background artifacts that appear as vertical striping in CNN-dominant models and mitigates the subtle checkerboard and phase misalignment visible in transformer/implicit baselines. The joint improvement in sharpness, canal continuity, and artifact suppression across both MIP views and 3D renderings supports the claim that the KAN lift with Koopman token dynamics and arch-aligned inverse warping yields perceptually superior, anatomy-consistent reconstructions from a single PX.

\subsubsection{Quantitative results}

\begin{table*}[htbp!]
	\centering
	\caption{Quantative comparison of the 3D oral reconstruction from single panoramic X-ray radiograph. The best results are highlighted in bold. The format is \textit{mean}$\pm$\textit{std} with $10$ repetitions of experiments.}
	\label{tab1}
	\begin{small}
	\renewcommand{\arraystretch}{1.2}%
	\begin{tabular}{lccccr}  
		\toprule
		Method &   Residual CNN  &  Oral-3D & 3DentAI & ViT- NeBLa & \textbf{K-U-KAN}(Ours) \\
		\midrule
		PSNR ($\uparrow$) &   $17.30 \pm 0.13$ &  $17.70 \pm 0.13$ & $20.05 \pm 0.40$ &  $\mathbf{23.4762 \pm 0.7823}$ & $23.4202 \pm 0.6863$\\
		SSIM(\%) ($\uparrow$)&  $35.01 \pm 04.03$ &  $36.05 \pm 04.20$ & $69.03 \pm 03.31$ &  $\mathbf{74.93 \pm 02.56}$ & $74.86 \pm 02.58$\\
		LPIPS (VGG) ($\downarrow$) &  $0.5157 \pm 0.0028$ &  $0.5143 \pm 0.0034$ & $0.4787 \pm 0.0082$ &  $0.4204 \pm 0.0093$ & $\mathbf{0.3212 \pm 0.0117}$\\
		Training Time ($\downarrow$) & \textbf{9 hr 44 min} &   11 hr 56 min & 9 hr 49 min & 22 hr 50 min & 10 hr 43 min\\
		\bottomrule
	\end{tabular}
	\end{small}
	\vskip -0.1in
\end{table*}

\begin{table*}[htbp!]
	\centering
	\caption{Comparision of perceptual similarity of the predicted CBCT volume from the models mentioned in the table. A lower value indicates higher perceptual similarity.}
	\label{tab:Lpips}
	\begin{small}
	\renewcommand{\arraystretch}{1.2}%
	\begin{tabular}{lccccr}  
		\toprule
		LPIPS ($\downarrow$)&  Residual CNN  & Oral-3D  & 3DentAI & ViT-NeBLa  & \textbf{K-U-KAN}(Ours)\\
		\midrule
		VGG Net &  $0.5157 \pm 0.0028$ &  $0.5143 \pm 0.0034$ & $0.4787 \pm 0.0082$ & $0.4204 \pm 0.0093$ & $\mathbf{0.3212 \pm 0.0117}$\\
		Alex Net &  $0.5804 \pm 0.0073$ &  $0.5855 \pm 0.0018$ & $0.3553 \pm 0.0136$ & $0.3285 \pm 0.0143$& $\mathbf{0.1772 \pm 0.0145}$\\
		Squeeze Net &  $0.3817 \pm 0.0055$ &  $0.3804 \pm 0.0028$ & $0.3210 \pm 0.0112$ & $0.2581 \pm 0.0133$& $\mathbf{0.1263 \pm 0.0107}$\\
		\bottomrule
	\end{tabular}
	\end{small}
	\vskip -0.1in
\end{table*}

We report fidelity (PSNR), structural agreement (SSIM), and perceptual similarity (LPIPS) on the held-out test split, averaging each score over ten independent runs (mean~$\pm$~std). Table~\ref{tab1} summarizes signal- and structure-level metrics, while Table~\ref{tab:Lpips} expands the LPIPS analysis across three feature backbones (VGG, AlexNet, SqueezeNet). These metrics, together with the training-time column, allow us to assess not only the numerical accuracy of each method but also the cost–quality trade-off that is crucial for clinical deployment (fast, stable models with strong perceptual realism are preferred).

\paragraph{\textbf{Signal and structure (PSNR/SSIM).}}
On Table~\ref{tab1}, \textbf{K--U--KAN} attains PSNR $23.4202\pm0.6863$~dB and SSIM $74.86\pm2.58$\%, essentially matching the strongest baseline, ViT--NeBLa ($23.4762\pm0.7823$~dB, $74.93\pm2.56$\%). Both substantially outperform CNN-only pipelines---Residual~CNN and Oral-3D remain near $17$--$18$~dB PSNR and mid-$30$\% SSIM, while 3DentAI improves structure ($69.03\pm3.31$\%) but remains below transformer/Koopman–KAN hybrids in absolute fidelity. This parity with ViT--NeBLa on PSNR/SSIM indicates that our \emph{learned lift $\rightarrow$ linear Koopman step $\rightarrow$ focal-trough warping} retains the high-frequency dental morphology and global jaw layout captured by transformer-driven implicit fields, but does so within a grid-efficient U-Net scaffold.

\paragraph{\textbf{Perceptual realism (LPIPS) and cross-backbone consistency.}}
The most salient gap appears in perceptual space. With VGG features, \textbf{K--U--KAN} achieves the best LPIPS ($\mathbf{0.3212\pm0.0117}$), improving markedly over ViT--NeBLa ($0.4204\pm0.0093$) and all other baselines (Table~\ref{tab1}). Table~\ref{tab:Lpips} shows this advantage is \emph{backbone-agnostic}: on AlexNet, \textbf{K--U--KAN} reaches $\mathbf{0.1772\pm0.0145}$ (vs.\ $0.3285\pm0.0143$ for ViT--NeBLa and $0.3553\pm0.0136$ for 3DentAI); on SqueezeNet, it achieves $\mathbf{0.1263\pm0.0107}$ (vs.\ $0.2581\pm0.0133$ for ViT--NeBLa and $0.3210\pm0.0112$ for 3DentAI). Because LPIPS correlates with human perception of sharpness and textural plausibility, these results suggest that the Koopman-regularized KAN lift preserves enamel–dentin interfaces, cortical bone edges, and mandibular canal courses more convincingly than the competing models, even when signal-level metrics are similar.

\paragraph{\textbf{Efficiency and practicality (training time).}}

Training-time comparisons in Table~\ref{tab1} highlight an important practical win. \textbf{K--U--KAN} completes in \emph{10~hr~43~min}, roughly \textbf{$\sim$2.1$\times$ faster} than ViT--NeBLa (22~hr~50~min), while delivering comparable PSNR/SSIM and \emph{substantially} better LPIPS. Although 3DentAI and Residual~CNN train slightly faster ($\approx$9~hr~45~min), they lag notably in perceptual quality and, for the CNNs, also in PSNR/SSIM. This places \textbf{K--U--KAN} on a favorable Pareto frontier: near–state-of-the-art signal/structure, state-of-the-art perceptual similarity, and materially reduced training cost relative to the strongest implicit/transformer baseline.

\begin{figure*}[htbp!]
	\centering
	\includegraphics[width=\linewidth]{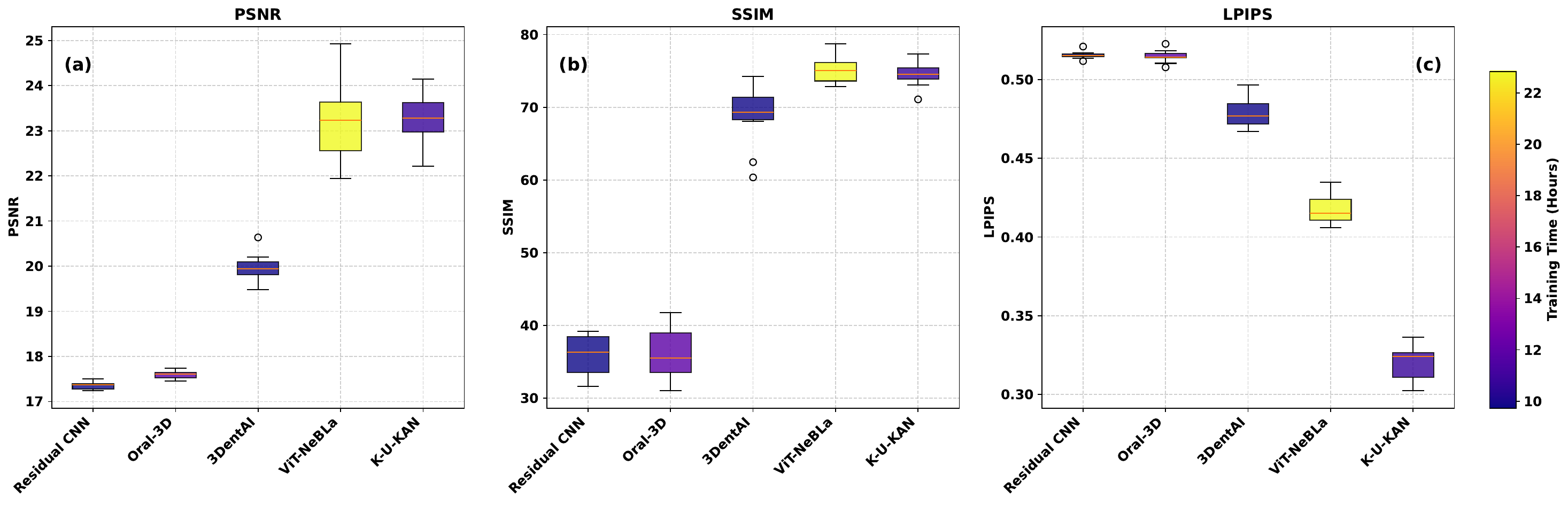}
	\caption{Boxplots of the (a) PSNR (dB) ($\uparrow$), (b) SSIM (\%) ($\uparrow$), and (c) LPIPS (VGG) ($\downarrow$) metrics visually summarize the performance stability of the five reconstruction methods evaluated: Residual CNN \cite{henzler_2018}, Oral-3D \cite{song_2021}, 3DentAI \cite{3dentai}, ViT-NeBLa \cite{vit_nebla_2025}, and K-U-KAN (Ours). The fill color of each box represents the training duration, providing a dual comparison of performance distribution and computational efficiency; darker hues indicate faster convergence while lighter hues (yellow) denote higher temporal cost. K-U-KAN achieves a high mean PSNR ($\sim 23.42$ dB) and SSIM ($\sim 74.86\%$) comparable to the state-of-the-art ViT-NeBLa, while securing the lowest mean LPIPS ($\sim 0.32$) and maintaining a training efficiency similar to lighter CNN baselines.}
	\label{fig:box_plots}
\end{figure*}

\paragraph{\textbf{Global Efficiency and Statistical Stability (Figs.~\ref{fig:bubble_plots} \& \ref{fig:box_plots}).}}To rigorously evaluate the trade-off between reconstruction quality and deployment feasibility, we visualize the relationship between performance metrics, training duration, and computational complexity in Fig.~\ref{fig:bubble_plots}. Here, the bubble size and color intensity scale with GFLOPS. The plots reveal that \textbf{K--U--KAN} achieves a superior efficiency-performance trade-off: it matches the high PSNR/SSIM and superior LPIPS of the transformer-based ViT-NeBLa, yet maintains a compact computational footprint (smaller, darker bubbles) comparable to the lightweight Residual CNN. Conversely, while ViT-NeBLa achieves strong metric peaks, it acts as a computational outlier with significantly higher GFLOPS and training latency.To further assess consistency across independent trials, Fig.~\ref{fig:box_plots} presents box-and-whisker plots for PSNR, SSIM, and LPIPS, with fill colors encoding the computational cost (training duration). In the PSNR and SSIM distributions (Fig.~\ref{fig:box_plots}(a-b)), \textbf{K--U--KAN} exhibits a tight interquartile range (IQR) centered around medians of $\approx$23.4~dB and $\approx$74.9\% respectively. This matches the stability profile of ViT--NeBLa while strictly dominating the spread of CNN baselines (Oral-3D, Residual CNN), which fluctuate at significantly lower performance tiers. Crucially, the LPIPS plot (Fig.~\ref{fig:box_plots}(c)) confirms that our perceptual advantage is systematic, with the entire distribution lying well below the best-performing baseline (ViT-NeBLa). Furthermore, the box color gradient reinforces the efficiency gap observed in the bubble plots: while ViT-NeBLa's high performance comes at the cost of the longest training duration (depicted in bright yellow), \textbf{K--U--KAN} maintains this high-fidelity, low-variance performance envelope with a much darker hue (purple/blue), indicative of a training time ($\sim$10 hours 43 min) comparable to the lightweight but lower-performing CNN architectures.

\subsubsection{Ablation study}

\begin{table*}[htbp!]
	\centering
		\caption{\textbf{Ablation on the refinement stage.} Replacing the attention 3D U–KAN refiner with a plain 3D U–Net yields similar PSNR/SSIM but noticeably worse LPIPS(VGG). The attention 3D U–KAN improves perceptual quality (sharper enamel–dentin and trabecular detail) without added cost, training slightly faster (10 hr 43 min vs.\ 10 hr 52 min).}
	\label{tab:ablation}
	
	\begin{small}
	\renewcommand{\arraystretch}{1.2}%
	\begin{tabular}{lcccr}  
        \toprule
		Methods &  PSNR($\uparrow$) & SSIM(\%)($\uparrow$) & LPIPS(VGG)($\downarrow$) & Training Time \\
		\midrule
		\textbf{K-U-KAN (w/ attention U-KAN refinement)} & $\mathbf{23.4202 \pm 0.6863}$& $\mathbf{74.86 \pm 02.58}$ & $\mathbf{0.3212 \pm 0.0117}$ & \textbf{10  hr 43 min}  \\
		K-U-KAN(w/ U-Net refinement)  & $23.3951 \pm 0.6649$ & $74.40 \pm 02.80$ & $0.3295 \pm 0.0118$ & 10 hr 52 min   \\
		\bottomrule
	\end{tabular}
	\end{small}
	\vskip -0.1in
\end{table*}

Table~\ref{tab:ablation} isolates the contribution of the \emph{attention 3D U--KAN} refiner by replacing it with a plain 3D U-Net while keeping the upstream K--U--KAN lift and inverse warping fixed. The full model with attention U--KAN achieves the best PSNR and SSIM ($23.4202 \pm 0.6863$~dB, $74.86 \pm 2.58$\%), with gains over the U-Net refiner that are modest and within one standard deviation ($+0.025$~dB PSNR, $+0.46$~SSIM points). The more decisive effect appears in perceptual space: LPIPS(VGG) drops from $0.3295 \pm 0.0118$ to $\mathbf{0.3212 \pm 0.0117}$ (absolute $\downarrow 0.0083$, $\sim$2.5\% relative), indicating crisper enamel–dentin transitions and more coherent trabecular textures for comparable signal-level metrics. Notably, this improvement comes with no cost penalty; training time is slightly lower for the attention U--KAN variant (10 hr 43 min vs.\ 10 hr 52 min). These trends support our design hypothesis: attention-gated skips suppress distractors in the encoder features, while the Tok–KAN bottleneck supplies expressive yet compact channel mixing—together yielding perceptual sharpness and anatomical continuity gains that do not rely on heavier computation.

\subsubsection{Model complexity and efficiency}

\begin{table*}[htbp!]
	\renewcommand{\arraystretch}{1.2}%
		\caption{\textbf{Model complexity and efficiency.} Comparison of trainable parameters (millions), static per-forward GFLOPS, and wall-clock training time under identical hardware and protocol. GFLOPS are computed with a FLOPs profiler by dividing the per-GPU forward operation count by the measured forward latency and expressing the resulting throughput in billions of operations per second. Despite a larger parameter budget, \textbf{K-U-KAN} with attention U-KAN refinement maintains a similar per-step compute cost and trains substantially faster than ViT-NeBLa \cite{vit_nebla_2025}, while CNN baselines show higher compute or longer time without matching quality. Training times are averaged over repeated runs.}
	\label{tab:complexity}
	\begin{center}
	\begin{small}
	\begin{tabular}{lccr} 
		\toprule  
		Methods &  Parameters (M) & GFLOPS &  Training Time \\
		\midrule
		\textbf{K-U-KAN (w/ attention U-KAN refinement)} & 151.512  & 127.36& 10 hr 43 min   \\
		K-U-KAN(w/ U-Net refinement)  & 131.504 & 128.72 & 10 hr 52 min    \\
		Vit-NeBLa \cite{vit_nebla_2025}& 59.814 & 90.47 & 22 hr 50 min  \\
		3DentAI \cite{3dentai} & 60.912 & 164.44 & 9 hr 49 min  \\
		Oral-3D \cite{song_2021}& 26.448  & 110.48 & 11 hr 56 min  \\
		Residual-CNN \cite{henzler_2018}& 43.298&200.73 & 9 hr 44 min   \\
		\bottomrule
	\end{tabular}
	\end{small}
	\end{center}
	\vskip -0.1in
\end{table*}


Table~\ref{tab:complexity} contrasts parameter count, floating–point workload, and wall–clock training time across methods. The full \textbf{K–U–KAN} with attention 3D U–KAN refinement carries 151.5M parameters and a static workload of 127.36~GFLOPs per forward pass, yet trains in \textbf{10~hr~43~min}, markedly faster than ViT–NeBLa (59.8M, 90.47~GFLOPs, \emph{22~hr~50~min}). This indicates that end–to–end efficiency is governed not only by parameter count or nominal FLOPs, but also by kernel fusion, memory locality, and the balance of convolutional vs.\ attention vs.\ token operations. Indeed, although 3DentAI and Residual–CNN exhibit higher nominal FLOPs (164.44 and 200.73~GFLOPs, respectively) and similar or slightly shorter times (\mbox{9~hr~49~min} and \mbox{9~hr~44~min}), their perceptual fidelity trails ours (cf.\ Table~\ref{tab:Lpips}), placing \textbf{K–U–KAN} a superior efficiency-performance trade-off (Fig. ~\ref{fig:bubble_plots}): near–minimal training time among high–fidelity models, with the best LPIPS by a large margin.

\paragraph{Throughput measurement.}
We compute computational efficiency with the \emph{DeepSpeed FLOPs Profiler}, which measures (per GPU) the forward floating-point operation count and the corresponding latency, then reports the realized throughput as FLOPs per second (FLOPS) \cite{deepspeed_2025}. Formally, letting
$F_{fwd}^{GPU}$ denote the per-GPU forward FLOP count and $\tau_{fwd}$ the measured forward latency (seconds), the per-GPU forward throughput is $\Phi_{fwd}^{GPU}= F_{fwd}^{GPU}/\tau_{fwd}$ (FLOPS) and GFLOPS = $\Phi_{fwd}^{GPU}/10^{9}$. We pair these runtime throughputs with the static per-forward GFLOPs in Table~\ref{tab:complexity} to expose how architectural choices translate into hardware utilization.

\section{Conclusions} \label{sec:conclusion}
We introduced K-U-KAN, a Koopman-enhanced U-KAN framework that reconstructs  3D oral anatomy from a single panoramic radiograph by combining three ingredients: a learned KAN lift that maps pixels to depth-aware observables; a Koopman token step that advances those observables via a contractive, phase-aware linear evolution (shrink + twist + bounded nudge); and an arch-aligned inverse warp that deposits predicted depth bins onto focal-trough rays, followed by a 3D attention U-KAN refiner. The design is rooted in the Beer–Lambert forward model, respects dental-arch geometry, and spectrally regularizes the ill-posed 2D $\to$ 3D lift through Koopman dynamics. 

Empirically, K-U-KAN achieves near state-of-the-art PSNR/SSIM, markedly better LPIPS, and substantially lower training time than transformer/implicit baselines such as ViT-NeBLa, indicating a strong balance among accuracy, perceptual realism, and computational efficiency. Ablations show that replacing the refiner with a plain 3D U-Net reduces perceptual sharpness despite similar signal-level scores, underscoring the value of tokenized KAN capacity and attention-gated skips for preserving enamel–dentin interfaces, cortical borders, and canal continuity. Qualitative comparisons corroborate these findings, with consistent suppression of background striping and phase misalignment artifacts and improved depiction of fine structures across coronal and sagittal MIPs.

Scientifically, the results support the premise that evolving measurements in learned coordinates—rather than directly regressing densities or relying on heavy positional encodings—offers a stable and interpretable route to volumetric recovery from a single PX. The complex-diagonal Koopman step serves as both a spectral brake and a phase carrier, enabling gentle, arch-following depth allocation that remains numerically well-behaved under native intensity scales and mixed precision. Practically, operating on grid features with hardware-efficient convolutions and concentrating KAN expressivity at token bottlenecks explains the observed runtime gains even at comparable parameter counts.

Limitations suggest clear next steps: (i) prospective clinical validation across vendors and acquisition protocols to assess robustness to positioning, metal artifacts, and pathology; (ii) uncertainty quantification (e.g., Bayesian heads or ensemble Koopman spectra) to support risk-aware decisions; (iii) weak supervision and geometry self-calibration to mitigate trajectory/pose mismatch; and (iv) coupling the Koopman-regularized lift with lightweight implicit decoders or focal-trough–aware sampling to further close the expressivity gap without sacrificing speed.

In summary, K-U-KAN advances radiation-efficient maxillofacial reconstruction by unifying physics, geometry, and dynamical-systems structure in a simple, stable token pipeline. 



\section*{Software and Data}

The datasets used in this study are not available publicly. Requests for dataset should be request to W. You and the implemented code materials should be addressed to B. K. Parida.

\section*{Acknowledgements}
We would like to thank Prof. Seong-Yong Moon from College of Dentistry in Chosun University  for facilitating the dataset. 
%
%
%
\section*{Authors Contributions}
\textbf{B.K. Parida}- Conceptualization of this study,
Methodology, Investigation, code implementation, experimentation, visualization, Writing - original draft, review
\& editing. \textbf{A. Sen - } writing - original draft, review \& editing. \textbf{W. You - } review \& editing,  funding acquisition, Supervison.

\section*{Conflict of Interest}
The authors declare no competing interests.

\nocite{langley00}

\bibliography{references}
\bibliographystyle{icml2026}

\newpage
\appendix

\section{Additional Figures} \label{app:additional_fig}
\begin{figure*}[htbp!]
	\centering
	\includegraphics[width=0.85\linewidth]{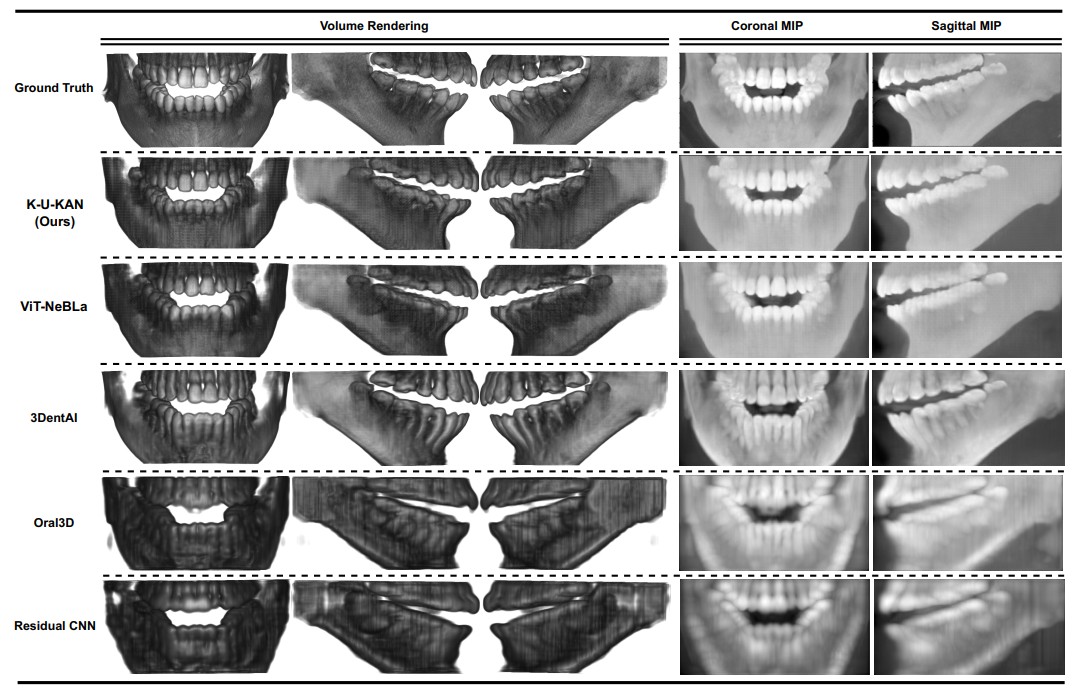}
	\caption[Qualitative comparison of reconstructed CBCT volumes]{Qualitative comparison of reconstructed CBCT volumes. From top to bottom: ground truth; K-U-KAN (ours), ViT-NeBLa \cite{vit_nebla_2025}; 3DentAI \cite{3dentai}; Oral-3D (auto-encoder) \cite{song_2021};  and residual CNN \cite{henzler_2018}. Columns show (left) volume renderings, (middle) coronal MIPs, and (right) sagittal MIPs.
	}
	\label{fig:qc_2_k}
\end{figure*}

\begin{figure*}[htbp!]
	\centering
	\includegraphics[width=0.85\linewidth]{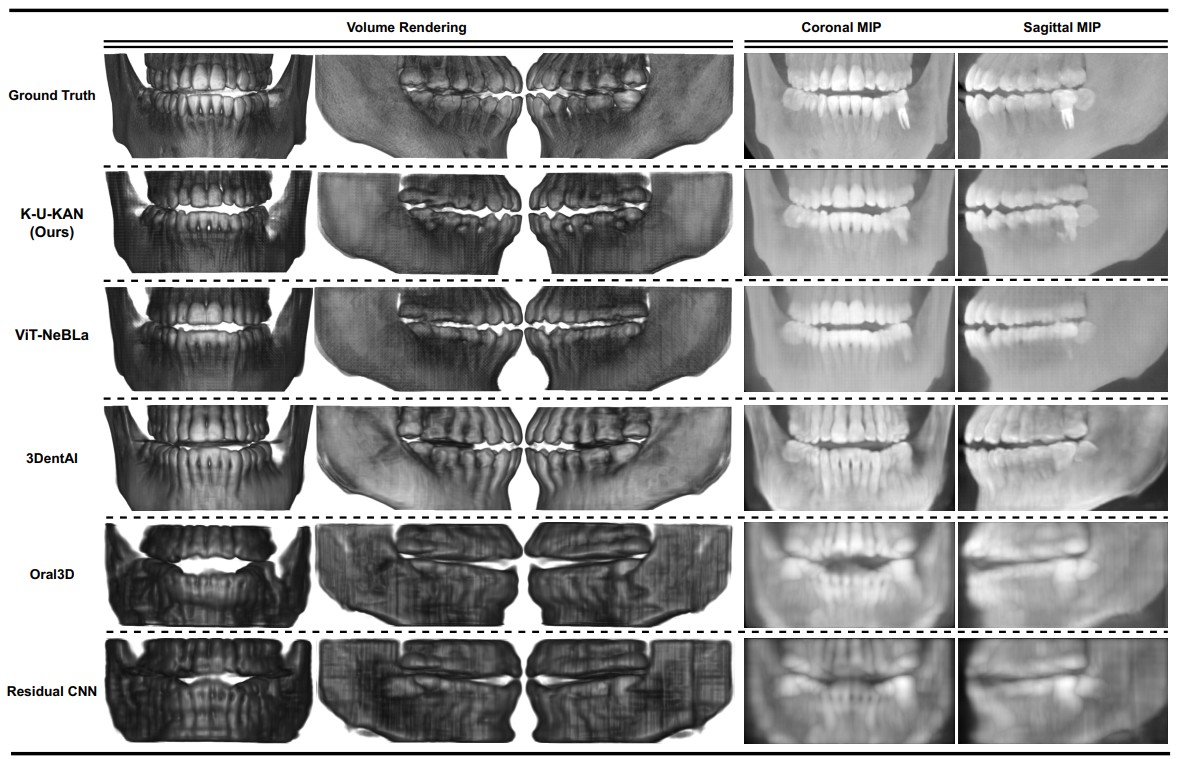}
	\caption[Qualitative comparison of reconstructed CBCT volumes]{Qualitative comparison of reconstructed CBCT volumes. From top to bottom: ground truth; K-U-KAN (ours), ViT-NeBLa \cite{vit_nebla_2025}; 3DentAI \cite{3dentai}; Oral-3D (auto-encoder) \cite{song_2021};  and residual CNN \cite{henzler_2018}. Columns show (left) volume renderings, (middle) coronal MIPs, and (right) sagittal MIPs.
	}
	\label{fig:qc_3_k}
\end{figure*}

\begin{figure*}[htbp!]
	\centering
	\includegraphics[width=0.85\linewidth]{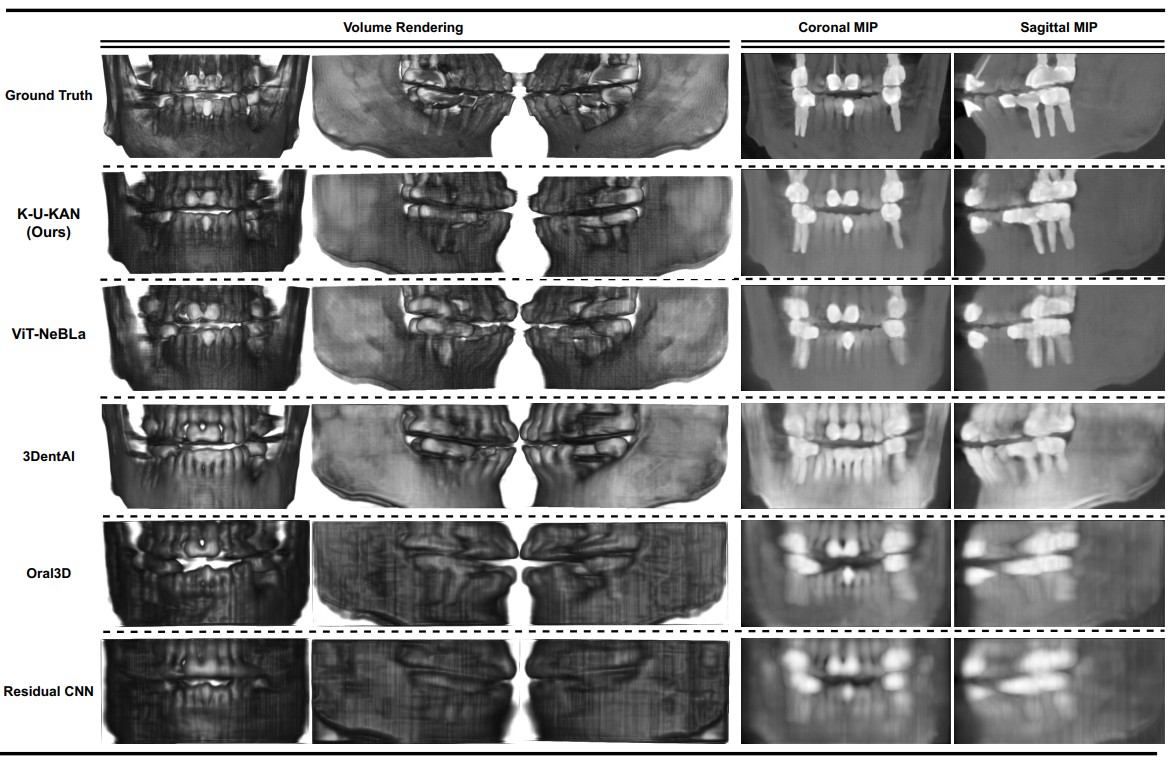}
	\caption[Qualitative comparison of reconstructed CBCT volumes]{Qualitative comparison of reconstructed CBCT volumes. From top to bottom: ground truth; K-U-KAN (ours), ViT-NeBLa \cite{vit_nebla_2025}; 3DentAI \cite{3dentai}; Oral-3D (auto-encoder) \cite{song_2021};  and residual CNN \cite{henzler_2018}. Columns show (left) volume renderings, (middle) coronal MIPs, and (right) sagittal MIPs.
	}
	\label{fig:qc_4_k}
\end{figure*}

\begin{figure*}[htbp!]
	\centering
	\includegraphics[width=0.85\linewidth]{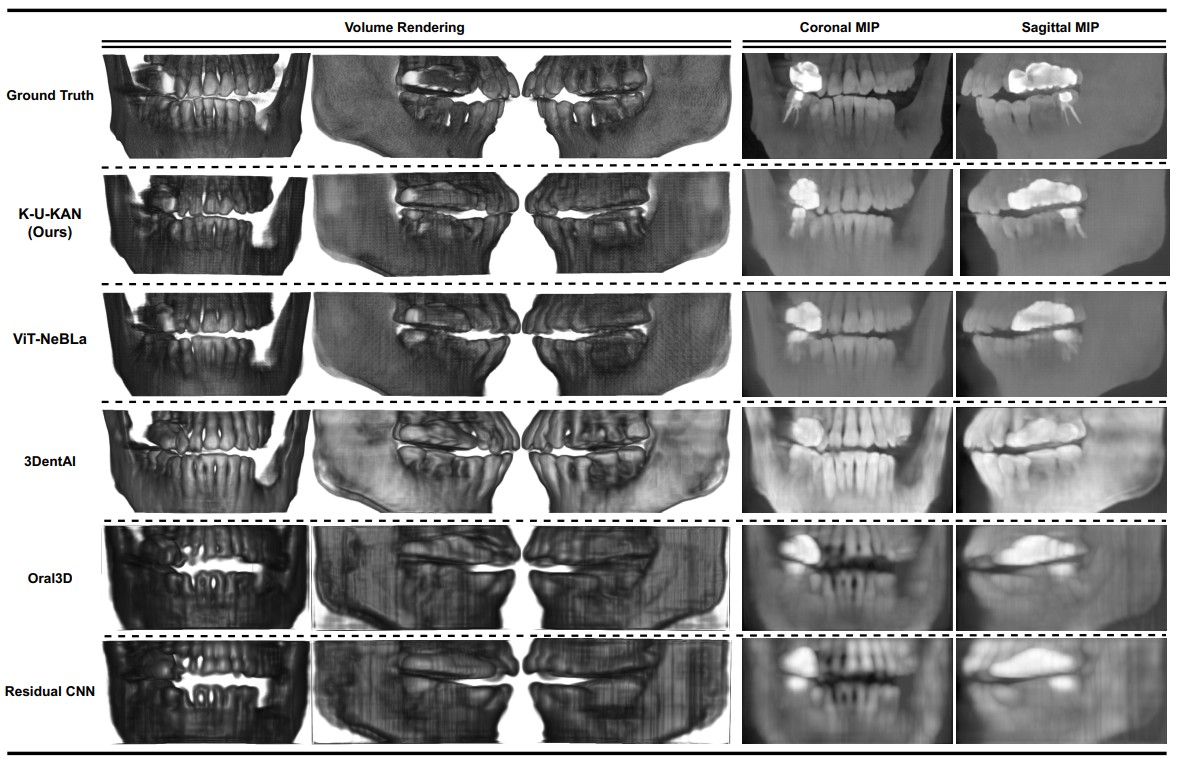}
	\caption[Qualitative comparison of reconstructed CBCT volumes]{Qualitative comparison of reconstructed CBCT volumes. From top to bottom: ground truth; K-U-KAN (ours), ViT-NeBLa \cite{vit_nebla_2025}; 3DentAI \cite{3dentai}; Oral-3D (auto-encoder) \cite{song_2021};  and residual CNN \cite{henzler_2018}. Columns show (left) volume renderings, (middle) coronal MIPs, and (right) sagittal MIPs.
	}
	\label{fig:qc_5_k}
\end{figure*}

In this section, additional qualitative reconstructions (Figs.~\ref{fig:qc_2_k}, \ref{fig:qc_3_k}, \ref{fig:qc_4_k}, \& \ref{fig:qc_5_k}) are reported to further assess the method’s robustness.
Each figure presents both volume‐rendered views and MIPs for distinct PX inputs, illustrating how our pipeline consistently recovers detailed 3D anatomy--preserving cortical surfaces, fine trabecular patterns, and dental landmarks--while suppressing background artifacts across varied patient cases.

\end{document}